\documentclass{article}

\usepackage{microtype}
\usepackage{graphicx}
\usepackage{subcaption}
\usepackage{booktabs} 

\usepackage{hyperref}

\usepackage[table]{xcolor}

\usepackage[accepted]{icml2026}

\usepackage{amsmath}
\usepackage{amssymb}
\usepackage{mathtools}
\usepackage{amsthm}
\usepackage{graphicx}
\usepackage{xcolor}
\usepackage{afterpage}

\usepackage{subcaption}

\newcounter{mycounter} 

\usepackage{tcolorbox}

\definecolor{fontblue}{RGB}{  0, 70,160}
\newcommand{\relativetext}[1]{{\small \color{fontblue}{#1}}}
\newcommand{\shadowtext}[2][yellow!20]{%
  \begingroup
  \setlength{\fboxsep}{0pt}
  \colorbox{#1}{#2}
  \endgroup
}

\newcommand{\shadowdowntext}[2][red!10]{%
  \begingroup
  \setlength{\fboxsep}{0pt}
  \colorbox{#1}{#2}
  \endgroup
}

\usepackage[capitalize,noabbrev]{cleveref}

\theoremstyle{plain}

\theoremstyle{definition}

\theoremstyle{remark}

\usepackage[textsize=tiny]{todonotes}

\icmltitlerunning{Flex-Forcing: Towards a Unified Autoregressive and Bidirectional Video Diffusion Model}

\begin{document}

\twocolumn[{
  \icmltitle{Flex-Forcing: Towards a Unified Autoregressive and Bidirectional \\ Video Diffusion Model}



  \icmlsetsymbol{equal}{*}

  \begin{icmlauthorlist}
    \icmlauthor{Xinyin Ma}{nus,nvidia}
    \icmlauthor{Julius Berner}{nvidia}
    \icmlauthor{Chao Liu}{nvidia}
    \icmlauthor{Arash Vahdat}{nvidia}
    \icmlauthor{Weili Nie}{nvidia,equal}
    \icmlauthor{Xinchao Wang}{nus,equal}
  \end{icmlauthorlist}
}

  \icmlaffiliation{nus}{Electrical and Computer Engineering, National University of Singapore, Singapore}
  \icmlaffiliation{nvidia}{NVIDIA Research, California, USA}

  \icmlcorrespondingauthor{Xinchao Wang}{xinchao@nus.edu.sg}
  \icmlcorrespondingauthor{Arash Vahdat}{avahdat@nvidia.com}

  \icmlkeywords{Machine Learning, ICML}

  \vskip 0.3in
]



\printAffiliationsAndNotice{\icmlEqualAdvising}

\begin{abstract}

Recent progress in large-scale generative models has substantially advanced video generation, yet existing methods remain constrained by a rigid inference paradigm. Bidirectional diffusion models excel at global coherence and visual fidelity but suffer from slow inference, while autoregressive models offer efficient and streaming generation at the cost of long-range consistency and exposure bias. We introduce Flex-Forcing, a unified training and inference framework that enables a video diffusion model to seamlessly operate under both bidirectional and autoregressive generation regimes. The core idea is a flexible chunking mechanism jointly defined over the temporal axis and denoising steps. This design allows the model to (1) perform flexible chunking according to different device budgets, (2) perform bidirectional inference across chunks for global structure planning, while generating frames autoregressively within each chunk for efficient and fine-grained synthesis, and (3) perform any-order, any-timestep autoregressive generation without the strict causal constraint. Extensive experiments on multiple video generation benchmarks demonstrate that Flex-Forcing achieves consistently better video quality, long-video stability than strong baselines with a rigid inference schedule, while offering faster inference. Project page: \url{https://research.nvidia.com/labs/genair/flex-forcing/}

\end{abstract}

\section{Introduction}

\begin{figure}[t]
    \centering
    \includegraphics[width=0.9\linewidth]{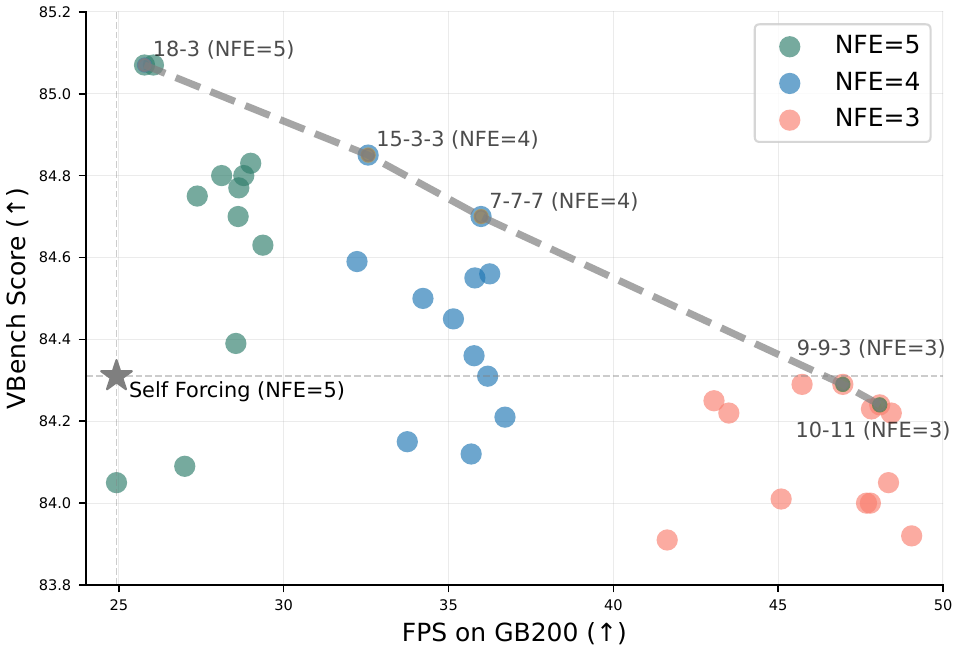}
    \caption{In various configurations, our method exhibits superior efficiency and performance compared to self-forcing, measured by FPS (frames per second) and VBench score. }
    \vspace{-3mm}
    \label{fig:pareto}
\end{figure}

Video generation has recently witnessed rapid progress driven by large-scale generative models \cite{openai2024sora,kuaishou2024kling,runway2024gen,google2024veo}, leading to substantial improvements in visual realism, temporal coherence and semantic consistency~\cite{yang2024cogvideox,wan2025wan,wu2025hunyuanvideo,chen2025seedance,hacohen2026ltx}. These advances have enabled a wide range of emerging applications, including long-form video synthesis~\cite{chen2025skyreels,yang2025longlive}, interactive world modeling~\cite{he2025matrix,hong2025relic}, and creative intelligence systems~\cite{hu2024animate,jiang2025vace}, where models are required to generate temporally extended, content-rich sequences with fine-grained control. As video models continue to scale in both capacity and context length, the efficiency and flexibility of their generative mechanisms have become increasingly critical, particularly for scenarios that demand customized and flexible trade-offs between quality, latency, and computational cost.

\begin{figure*}[t]
    \centering
    \includegraphics[width=\linewidth]{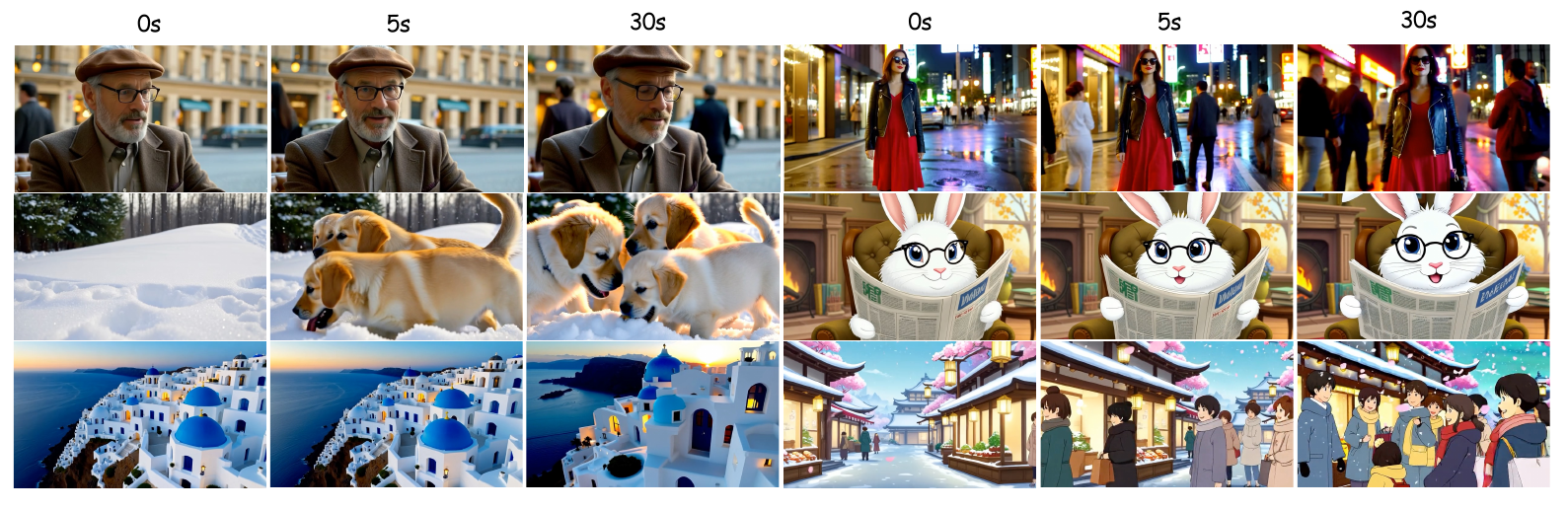}
    \vspace{-6mm}
    \caption{Generated examples from Flex-Forcing. We show three frames from the 30s 480p video under six different prompts.}
    \vspace{-3mm}
    \label{fig:example}
\end{figure*}

Among existing video generation approaches, two dominant paradigms have emerged: the bidirectional diffusion paradigm, widely adopted in pretrained video diffusion models \cite{ho2022video,hong2022cogvideo,chen2024videocrafter2}, and the autoregressive paradigm, which has recently emerged as an efficient alternative \cite{deng2024nove,wang2024loong,teng2025magi,yin2025causvid,huang2025selfforcing}, each offering complementary advantages. The bidirectional paradigm leverages full-context attention to jointly model all frames, resulting in strong temporal coherence and high visual fidelity \cite{ho2022video}, and is therefore well-suited for capturing long-range dependencies such as camera motion, scene transitions, and complex interactions \cite{yin2023nuwa,singer2022make}. However, the bidirectional paradigm involves substantial computational overhead, leading to high inference costs, which limits its applicability to real-time settings and its scalability to long video generation \cite{yin2025causvid,yu2023magvit}.
In contrast, autoregressive video generation models operate under causal conditioning, generating frames sequentially while reusing past key–value states via KV caching. This enables real-time inference and naturally supports arbitrary-length video generation without reprocessing earlier frames~\cite{hong2022cogvideo,villegas2022phenaki}. Nevertheless, the lack of global context during generation makes autoregressive models susceptible to error accumulation over time, resulting in drift in object appearance, motion, or scene structure, as well as weaker global temporal consistency and limited long-horizon planning ability \cite{liu2025rolling}.

Despite their complementary strengths, existing work lacks an effective mechanism to jointly realize the advantages of bidirectional and autoregressive video generation paradigms within a unified framework, and consequently optimizes for one paradigm while compromising the other \cite{yin2025causvid,huang2025selfforcing}. We therefore seek to unify both paradigms within a single model, enabling coherent video generation while supporting efficient long-video inference without increasing exposure bias or computational cost. To this end, we propose Flex-Forcing, a unified framework that enables test-time control over the inference paradigm. A single trained model can flexibly operate in (1) a bidirectional mode for globally consistent generation, (2) an autoregressive mode for scalable long-video or streaming synthesis, or (3) a semi-autoregressive hybrid mode that provides intermediate trade-offs between quality and efficiency under different deployment constraints.

To enable such a flexible inference behavior, the core challenge in training is to equip a single model with both causal (autoregressive) and non-causal (bidirectional) generation capabilities. This allows the model to operate in intermediate regimes between strictly causal and fully bidirectional inference.
We address this challenge by introducing \textit{Flexible Chunking}, defined along two orthogonal axes, temporal frames and denoising steps, under which autoregressive and bidirectional inference emerge as two extreme cases.
This formulation is consistently applied into both training and inference, 
which exposes the model to mixed causal and non-causal conditioning contexts. The same query token may attend to key-value pairs of different noise levels, where causal past tokens are typically more denoised, while non-causal future tokens are noisier.
To reconcile this noise-level discrepancy, we introduce an alignment mechanism that explicitly enforces representation consistency between causal and non-causal attention contexts.

Our results demonstrate that bidirectional and autoregressive inference can be unified within a single Flex-Forcing model, rather than treated as mutually exclusive paradigms. The resulting flexible inference mode achieves a strong quality–efficiency trade-off, yielding a markedly improved Pareto frontier (Figure~\ref{fig:pareto}) and substantially outperforming Self-Forcing~\cite{huang2025selfforcing} across both short- and long-video benchmarks.
Beyond generation, this flexible paradigm enables downstream applications with adaptive causal constraints, such as \textit{any-order, any-step autoregressive editing}, which supports localized temporal edits while preserving global video coherence with the original video.

\begin{figure*}[t]
    \centering
    \includegraphics[width=\linewidth]{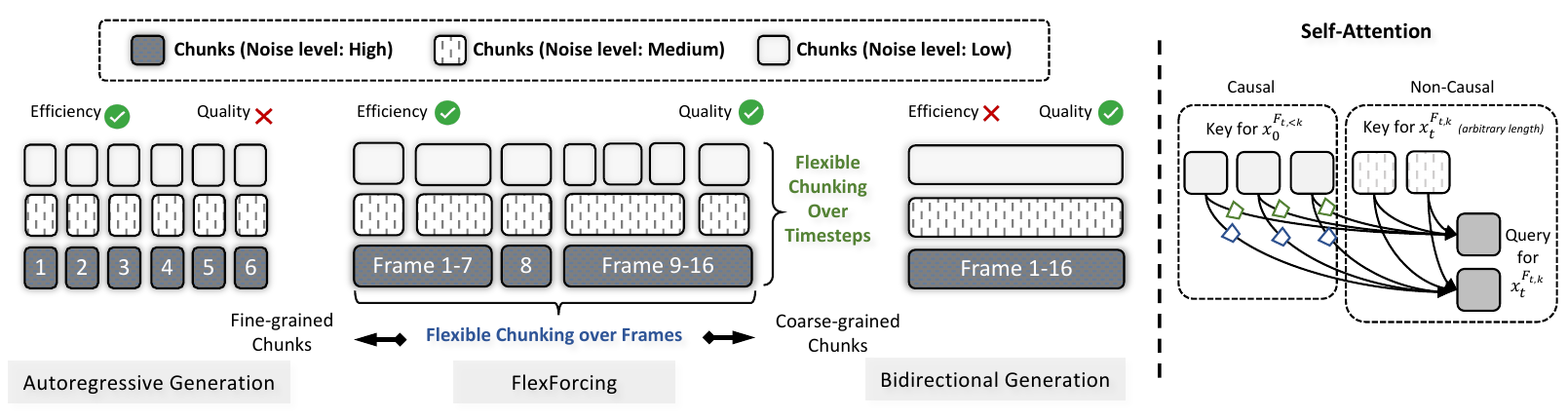}
    \caption{(Left) Flexible chunking for bridging the autoregressive and bidirectional video generation. Flex-Forcing adjusts chunk granularity across noise levels while a unified self-attention mechanism supports both causal and bidirectional inference. (Right) The mixed attention with causal tokens and non-causal tokens. We add a timestep dependent K-Projection at the clean cache from past frames.}
    \label{fig:main_illustration}
    \vspace{-4mm}
\end{figure*}

To summarize, our contributions are: 
\begin{itemize}
    \item We propose \textit{Flex-Forcing}, a unified framework that supports both bidirectional and autoregressive video generation within a single model.
    \item We introduce \textit{flexible chunking} along the temporal and denoising axes, together with a \textit{train--test consistent objective}, and an \textit{aligned conditioning} mechanism to bridge different causal contexts.
    \item Extensive experiments show that Flex-Forcing 
    substantially outperforms causal models on the performance-efficiency trade-off while achieving competitive performance with bidirectional models.
    \item We further introduce \textit{any-order, any-step autoregressive editing} based on Flex-Forcing, which enables controllable autoregressive editing of videos. 
\end{itemize}

\section{Related Work}

\paragraph{Bidirectional models for video generation.}
Early diffusion-based video generation models extend image diffusion models \cite{dhariwal2021diffusion,rombach2022high,karras2022elucidating,black2024flux,esser2024scaling} to the spatio-temporal domain by jointly denoising all frames with full temporal context \cite{ho2022video,bar2024lumiere,luo2023videofusion}. These models typically adopt bidirectional self-attention over the temporal axis, allowing each frame to attend to both past and future frames during denoising \cite{ho2022imagen}. Subsequent works scale this paradigm using transformer-based backbones and large-scale video–text datasets, establishing bidirectional diffusion models as foundation architectures for general-purpose text-to-video generation \cite{yang2024cogvideox,blattmann2023align,blattmann2023stable,gupta2024photorealistic}.

\paragraph{Autoregressive models for video generation.} 
Autoregressive video generation models generate videos by sequentially predicting frames, enabling real-time generation, prompt control and long video generation \cite{teng2025magi}. VideoGPT \cite{yan2021videogpt} uses a GPT-like model to autoregressively model  discrete latents using spatio-temporal position encodings. Pyramidal flow \cite{jin2024pyramidal} crafts autoregressive video generation with a temporal pyramid to compress the full-resolution history . Lumos-1 \cite{yuan2025lumos} introduces autoregressive discrete diffusion forcing to mitigate the frame-wise imbalance loss. Besides training native autoregressive video generation models, another line of work explores causal distillation \cite{yin2025causvid,huang2025selfforcing,cui2025selfforcing++,liu2025rolling}, which use diffusion forcing \cite{chen2024diffusion} and causal attention mask to initialize the model and then use DMD \cite{yin2024dmd,yin2024dmd2} and self-rollout \cite{huang2025selfforcing} in training.

\section{Flex-Forcing}

\paragraph{Overview.}
We propose a unified generative framework that supports both bidirectional and autoregressive inference. To the best of our knowledge, this is the first approach to demonstrate that these two paradigms can inherently coexist within a single trained model. 
The core idea is a flexible chunking strategy that composes chunks of varying granularity. Flexible chunking is learned by post-training a bidirectional video diffusion model and is defined along two orthogonal axes:
(i) the video-frame axis, which allows heterogeneous chunk sizes across different temporal regions; and
(ii) the denoising-timestep axis, which allows the chunk granularity to change throughout the denoising trajectory.
Together, they induce a rich configuration space and enable the model to seamlessly operate in bidirectional or autoregressive modes for quality–efficiency trade-offs.

\subsection{Flexible Chunking over Video Frames}

We represent a video as a sequence of $F$ frames
$x = (x^{(1)}, x^{(2)}, \ldots, x^{(F)})$. 
Diffusion sampling performs $T$ denoising steps with $\{x_t\}_{t=1}^{T}$.
A chunk consists of a consecutive sequence of frames and we make those chunks to be flexible in size. 
In this way, we obtain a hybrid factorization in which intra-chunk dependencies are modeled bidirectionally, while inter-chunk generation remains autoregressive. The partition is allowed to vary across denoising timesteps, enabling the factorization to adapt to the uncertainty of the generation process.
Specifically, at each denoising step $t$, we define a contiguous partition of frame indices $\{1,\ldots,F\}$ by chunk boundary indices $\mathbf{a}_t = (a_{t,0}, a_{t,1}, \ldots, a_{t,K_t})$, where $1=a_{t,0}<\cdots<a_{t,K_t}=F+1$. 
The $k$-th chunk at timestep $t$ covers the frame index set:
\begin{equation}
\mathcal{F}_{t,k} \triangleq \{\, f \mid a_{t,k-1} \le f < a_{t,k} \,\}, 
\end{equation}
At each denoising step $t$, given a partition $\mathbf{a}_t$, the chunk-wise causal sampling now is defined as:
\begin{equation}
x_{t-1}^{\mathcal{F}_{t,k}} \sim
q_\theta\!\Big(x_{t-1} ^{\mathcal{F}_{t,k}} \ \Big|\ x_{0}^{\mathcal{F}_{t,<k}},\ x_t^{\mathcal{F}_{t,k}};\mathbf{a}_t\Big),
\label{eqn:semi_ar}
\end{equation}
where $\mathcal{F}_{t,<k}\triangleq \bigcup_{u<k}\mathcal{F}_{t,u}$ denotes all frame indices before current chunk, and $x^{\mathcal{F}}$ denotes the frames in chunk $\mathcal{F}$. 
Following self-forcing \cite{huang2025selfforcing}, $x_{0}^{\mathcal{F}_{t,<k}}$ is represented by the KV cache of previously predicted frames in the diffusion transformer~\cite{peebles2023scalable}. 

Eq.~\ref{eqn:semi_ar} formalizes that later chunks are generated via denoising conditioned on earlier chunks.
The above formulation implies a flexible chunking strategy along the video's frame axis. By setting the frame-selection parameter $\mathbf{a}_t$, our framework defines a variable subset of frames that can act as the history under the causal constraint and the context that interacts in the bidirectinal attention. 
Table~\ref{tab:hybrid} provides a unified view of auoregressive and bidirectional inference as specific cases of our hybrid inference.

\renewcommand{\arraystretch}{1.235}
\begin{table}[htbp!]
    \centering
    \resizebox{\linewidth}{!}{
    \begin{tabular}{c|c|c}
    \toprule
        Mode & $ \mathbf{a}_t $ & $q_\theta(x_{t-1} \mid x_t)$\\
    \midrule
        Autoregressive & $(1, 2, 3, \ldots, F, F+1)$ & $\prod_{k=1}^{F} q_\theta\!\left( x_{t-1}^{k}\;\middle|\;x_0^{<k},\, x_t^k\right)$ \\
        Bidirectional & $(1, F+1)$ & $q_\theta(x_{t-1}^{1:F} \mid x_t^{1:F})$ \\
        Hybrid & $(a_{t,0}, a_{t,1}, \ldots, a_{t,K_t})$ & $\prod_{k=1}^{K_t} q_\theta\!\Big(x_{t-1}^{\mathcal{F}_{t,k}} \ \Big|\ x_{0}^{\mathcal{F}_{t,<k}},\ x_t^{\mathcal{F}_{t,k}};\mathbf{a}_t\Big)$\\
         \bottomrule
    \end{tabular}
    }
    \caption{Unified view of autoregressive, bidirectional, and hybrid inference as specific cases of the boundary configuration $\mathbf{a}_t$.}
    \label{tab:hybrid}
\end{table}

\subsection{Flexible Chunking over Denoising Timesteps.}
As defined in Eq.~\ref{eqn:semi_ar}, the  $t$-dependent chunk partition $\mathbf{a}_t$ allows the chunking strategy to vary across denoising timesteps.
Intuitively, early denoising steps, characterized by high noise levels, focus on global structure and benefit from larger temporal chunks. Conversely, later refinement steps prioritize local details, which can be effectively modeled with smaller chunks that require less long-range context. This results in a hierarchical, pyramid-like structure where chunk size decreases as noise levels drop (Figure~\ref{fig:main_illustration}).

Formally, recall that $\mathbf{a}_t=(a_{t,0},a_{t,1},\ldots,a_{t,K_t})$ denotes the frame partition at denoising timestep $t$.
We impose a \emph{nested flexibility} that allows each chunk to be further subdivided as denoising proceeds.
Specifically, after completing the denoising at step $t+1$ (i.e., $x_{t+1} \!\rightarrow x_{t}$), we split the partition for $t$ by \emph{inserting} additional boundaries while preserving all existing ones.
Timestep $t$ inherits the same configuration of $t+1$: $\mathcal{F}_{t,k}=[a_{t,k-1},a_{t,k}) = [a_{t+1,k-1},a_{t+1,k})$, and we introduce a sequence of new split points $\mathcal{S}_{t,k}$
with the number of sub-chunk boundaries $n_{t,k}\ge 0$. Simply merging the original endpoints and the split points gives the new chunking patterns:
\begin{equation}
\mathbf{a}'_{t,k}
\;=\;
a_{t,k-1} \cup \mathcal{S}_{t, k} \cup  a_{t,k},
\label{eq:refined_boundaries}
\end{equation}
This speratability can be applied \emph{recursively} and the resulting sub-chunks can be further split by inserting additional boundaries in the next denoising step. 

During inference, frames are processed in temporal order. However, under the pyramid chunking strategy, a synchronization challenge arises when a large chunk at denoising step $t$ is partitioned into smaller sub-chunks at step $t-1$. In these instances, we temporarily buffer the denoising results for all frames within the original bidirectional chunk at step $t$. We then autoregressively resume denoising for each sub-chunk at step $t-1$ once the required KV caches from preceding sub-chunks become available. This execution order ensures that causal dependencies are satisfied across varying granularities (see Figure \ref{fig:chunking} in Appendix).

\subsection{Flexible-chunk Training}

With the unified formulation in place, we next incorporate it into the training objective. The training consists of two parts:
(1) introducing causality constraints into a bidirectional diffusion model while preserving its non-causality ability to attend to future chunks, and
(2) enabling the model to attend to key states derived from inputs at different noise levels.

\paragraph{Injecting the causality when maintaining the non-casuality}
We largely follow the training paradigm introduced in CausVid~\cite{yin2025causvid} and Self-Forcing~\cite{huang2025selfforcing}, which transfers a bidirectional diffusion model into a causal one. This training pipeline consists of two stages.
(1) ODE initialization, where a causal attention mask is applied to inject causality into the model.
(2) Asymmetric distillation with DMD training, which further strengthens causality through self-rollout and applies the VSD loss~\cite{yin2024dmd} to distill a multi-step diffusion model into a few-step one.
We find that injecting causality while preserving non-causal modeling capacity can be naturally achieved during the asymmetric distillation stage.

In the asymmetric distillation stage, we introduce a stochastic chunking strategy that dynamically varies the attention pattern within each rollout.
Recall that we define a contiguous partition of the frame index set via a boundary configuration, we randomly sample a chunk partition $\mathbf{a}_t$ during training, and apply rollout to these flexible chunks in a dynamic way.
After the latents of all frames $\textbf{x}_0 =\{x_0^i\}_{i=1}^{F}$ has been rolled out under each $\mathbf{a}_t$, we train the generator $G_\theta$ by taking the gradient from~\cite{yin2024dmd}:
\begin{equation}
    \! \nabla_\theta D_{K L}=\underset{\substack{\mathbf{z} \sim \mathcal{N}(0 ; \mathbf{I})}}{\mathbb{E}}\left[-\left(s_{\text {real }}(\mathbf{x}_t)-s_{\text {fake }}(\mathbf{x}_t)\right) \frac{\partial G_\theta}{\partial \theta}\right]
\end{equation}
where $s_{\text {real }}$ and $s_{\text {fake}}$ denote the real and fake scores, respectively, and $\mathbf{x}_t = I(G_\theta(\mathbf{z}), t)$ with $I(\mathbf{x}_0, t)$ representing a forward process in diffusion models.
By randomly sampling the chunk size, the model is exposed to attention configurations that range from strictly causal to fully non-causal.
As a result, the training implicitly contains a heterogeneous mixture of causal and bidirectinal dependencies.

\paragraph{Noise-level alignment across causal and non-causal attention.} 
Since our model supports both causal and bidirectional inference, the same query token may attend to contexts produced under different generative regimes. At a given self-attention layer, this implies that a query can attend to key–value states derived from more denoised representations (causal past tokens) as well as states generated from noisier latents at higher diffusion timesteps (non-causal future tokens). As a result, the attended KV states exhibit substantially different signal-to-noise ratios. Standard self-attention treats all key–value pairs uniformly, which introduces a noise-level mismatch in context aggregation and degrades performance in flexible inference settings.

To address this issue, we propose a noise-aligned projection for the key states, termed K-Projection. Concretely, it projects the cache for key states generated from clean inputs into the ``noisy'' latent space corresponding to the current diffusion timestep. After projection, all key states are expressed in a noise-consistent representation space, enabling standard self-attention to be applied. 
Formally, the timestep-dependent projection is defined as:
\begin{align}
    & \Pi_{t \leftarrow 0}:\mathbb{R}^{d}\rightarrow\mathbb{R}^{d},\qquad
\tilde{K}_{t}=\Pi_{t \leftarrow 0}\!\left(K_{0}\right).
\end{align}
where $\Pi_{t \leftarrow 0}$ is a lightweight, timestep-conditioned linear projection and is intialized as identity mappings. For each denoising step $t$, the self-attention computation is then performed as:
\begin{align}
    & \operatorname{Attn}(Q_t, \tilde{K}_{t}, V_t) = \operatorname{softmax}\left(\frac{{Q_t} \tilde{K}_t^{T}}{\sqrt{d}}\right) V_t \nonumber \\
    & \text{where } 
    \tilde{K}_t = \operatorname{concat}\left(\Pi_{t \leftarrow 0}\!\left(K_{0}^{<\mathcal{F}_{t,k}}\right), \tilde{K}_t^{\mathcal{F}_{t,k}}\right) 
\end{align}
The projection is applied on-the-fly during attention computation, without modifying the cached KV tensors or gradient propogation on the KV cache. During inference, clean KV states are stored once and dynamically projected according to the diffusion timestep.
This design preserves the efficiency benefits of KV caching while enabling stable flexible inference across causal and non-causal regimes.

\begin{figure}[t]
    \centering
    \includegraphics[width=\linewidth]{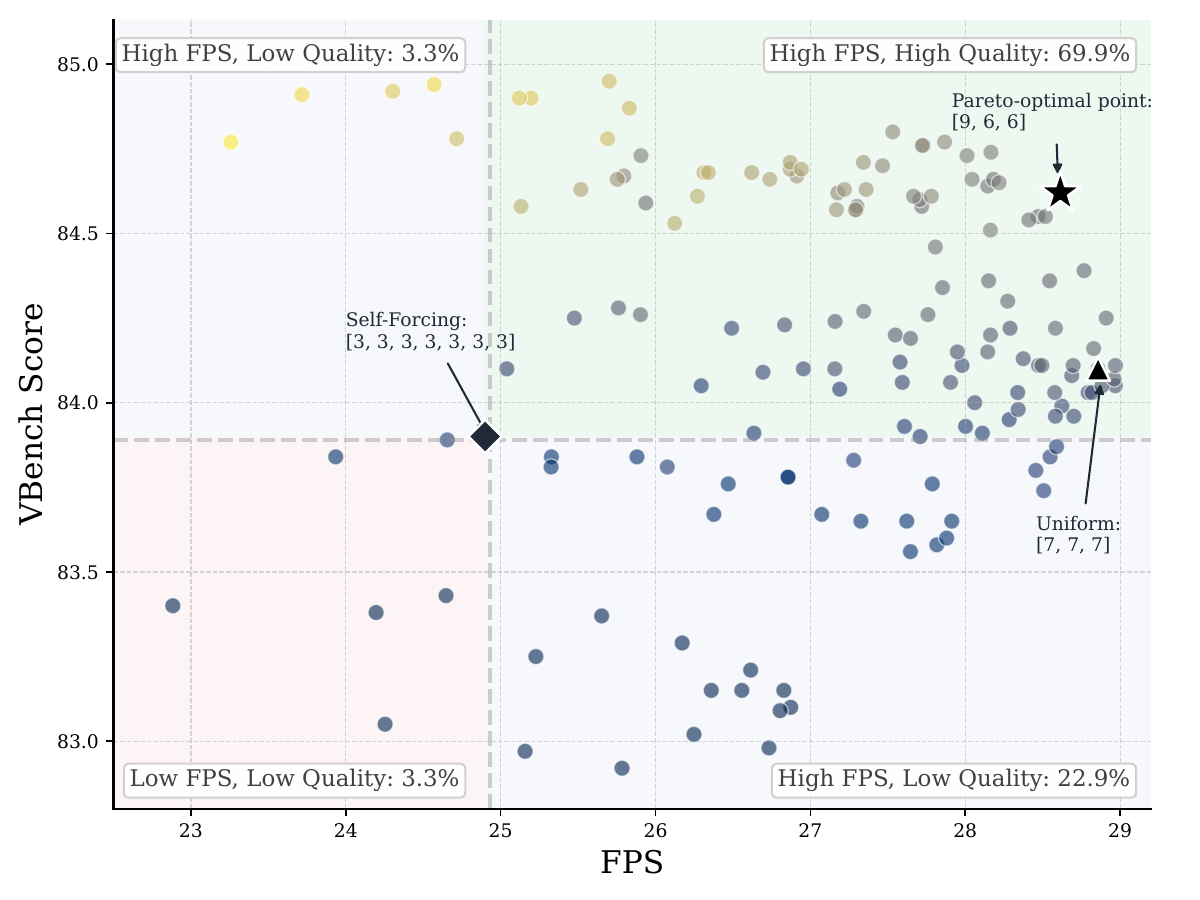}
    \caption{We test the FPS and VBench Score that split 5s videos into 3 chunks. Point color indicates the average temporal starting position of chunks, from earlier (blue) to later (yellow). 
    }
    \label{fig:brute-force-5s}
\end{figure}

\begin{figure*}[t]
    \centering
    \includegraphics[width=\linewidth]{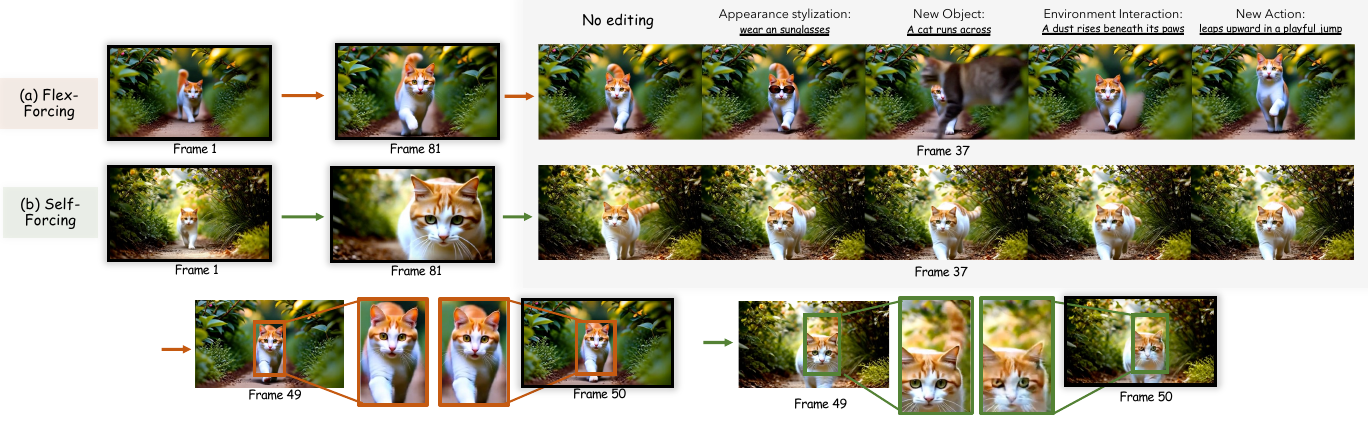}
    \caption{Autoregressive any-order generation under the Flex-Forcing paradigm. We first generate the full sequence of 81 frames and then re-edit an arbitrary temporal segment (frames 26 – 49). Our method achieves a high success rate for editing while preserving significantly better consistency with the unedited subsequent chunks (see the transition between Frame 49 and Frame 50).}
    \label{fig:any-order}
\end{figure*}

\section{Applications of Flex-Forcing}

\subsection{Inference Flexibility: Better Speed, Better Quality}
We begin by presenting the most straightforward application of Flex-Forcing, which enables a hybrid and flexible inference paradigm that combines autoregressive and bidirectional inference at test time. By allowing variable chunk sizes, Flex-Forcing supports adaptive and searchable chunk configurations that can be tailored to diverse compute budgets and videos of different lengths, and offers a more favorable trade-off between quality and efficiency.

We perform a brute-force search to identify the optimal chunk configuration for a 5-second video under a fixed constraint of three chunks. Concretely, we enumerate all valid partitions of the 21 latent frames into three chunks, excluding cases that a chunk only contain one frame. The results are shown in Figure~\ref{fig:brute-force-5s}, from which we draw the following key observations.

\begin{itemize}
\item Uniform chunking is not optimal. Evenly partitioning frames can lead to markedly inferior performance compared to asymmetric alternatives.
\item Chunk layout matters beyond exposure bias. Under the same number of exposure rounds, different chunk configurations exhibit large performance disparities.
\item Performance favors temporally front-loaded chunking. Allocating larger chunks to early frames and smaller chunks to later frames yields the strongest results, occasionally surpassing bidirectional inference.
\end{itemize}

\subsection{Autoregressive Any-timestep, Any-order Editing}

\begin{figure}[t]
    \centering
    \includegraphics[width=\linewidth]{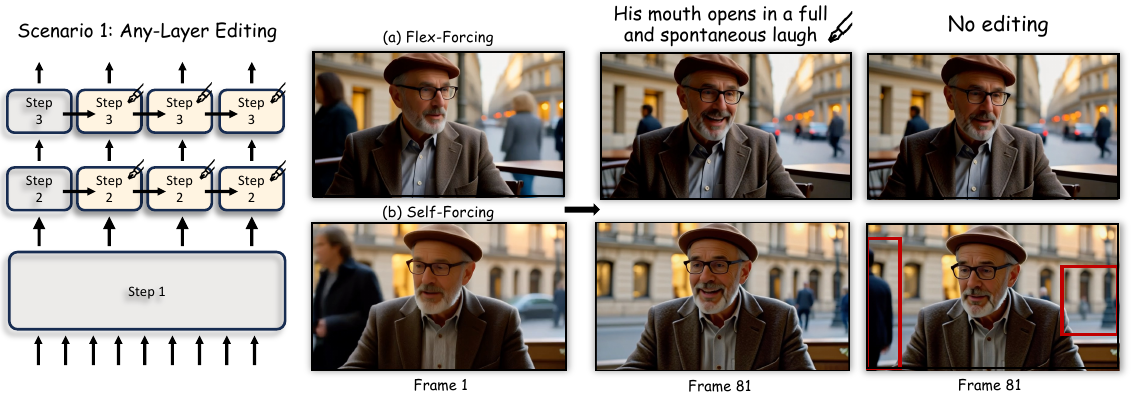}
    \caption{Autoregressive any-timestep editing. We modify the conditioning starting from the 6th frame during the last three of four denoising steps. For a fair comparison, the baseline applies the same conditioning updates at the identical frames and timesteps.
    }
    \label{fig:any-layer}
\end{figure}

Under Flex-Forcing, we unlock two new forms of editing that are difficult under conventional autoregressive paradigm. Specifically, we explore two directions: autoregressive any-timestep and any-order editing.

\paragraph{Autoregressive any-timestep editing.} 
To maintain global coherence, we adopt a structure-preserving editing strategy that restricts editing to low-level refinement timesteps while keeping high-level planning timesteps unchanged. This separation explicitly decouples global structure formation from local detail modification, resulting in substantially improved consistency over baseline autoregressive editing methods (Figure~\ref{fig:any-layer}).
Under the Self-Forcing paradigm, the globally coupled generation dynamics cause even small local edits to propagate across frames, often leading to large deviations in the final outputs.

\paragraph{Autoregressive any-order editing.}

Leveraging the ability to attend to both past and future tokens, our model enables any-order editing while still operating within an autoregressive generation framework. Editing a target chunk can thus be performed independently of its temporal order.
Formally, editing at timestep $t$ and chunk $k$ is defined as:
\begin{equation}
    x_{t-1}^{\mathcal{F}_{t, k}} \sim q_\theta\left(x_{t-1}^{\mathcal{F}_{t, k}} \mid x_0^{<\mathcal{F}_{t,k}}, x_0^{>\mathcal{F}_{t,k}}, x_{t}^{\mathcal{F}_{t, k}}\right)
\end{equation}
where $<\mathcal{F}_{t,k}$ and $>\mathcal{F}_{t,k}$ represents clean tokens belonging to chunks before and after $\mathcal{F}_{t,k}$. Conditioning on both sides allows arbitrary temporal segments to be re-edited after full-video generation without re-generating the entire sequence, relaxing the strict causal constraint of standard autoregressive models (Figure~\ref{fig:any-order}).

\begin{table}[t]
    \centering
    \resizebox{\linewidth}{!}{%
    \begin{tabular}{l|cccccc}
    \toprule
    Model & \#Params & FPS & NFE & \multicolumn{3}{c}{Evaluation Scores} \\
    & & & & Total & Quality & Semantic \\
    \midrule
    \rowcolor{gray!20} \multicolumn{7}{l}{\textit{Pre-trained Autoregressive / Diffusion Video Generation Models}}\\
    SkyReels-V2 \cite{chen2025skyreels}& 1.3B & 2.2 & 30$\times$2 & 82.67 & 84.70 & 74.53 \\
    MAGI-1 \cite{teng2025magi}     & 4.5B & 0.5 & 64$\times$3 & 79.18 & 82.04 & 67.74\\
    NOVA  \cite{deng2024nove}      & 0.6B & 0.9 & 25$\times$2 & 80.12 & 80.39 & 79.05 \\
    Pyramid Flow \cite{jin2024pyramidal} & 2B  & 0.9 & 20$\times$2 & 81.72 & 84.74 & 69.62 \\
    Wan2.1 ~\cite{wan2025wan}     & 1.3B &  1.3 & 50$\times$2 & 84.26 & 85.30 & 80.09 \\
    \midrule
    \rowcolor{gray!20} \multicolumn{7}{l}{\textit{Autoregressive Model Distilled from Diffusion Models}}\\
    CausVid ~\cite{yin2025causvid}    & 1.3B & 24.9 & 5 & 81.20 & 84.05 & 69.80\\
    Self Forcing (Chunk-wise) & 1.3B & 24.9 & 5 & 84.31 & 85.07 & 81.28 \\
    Self Forcing (Chunk-wise)* & 1.3B & 24.9 & 5 & 83.89 & 84.45 & \textbf{81.63} \\
    Self Forcing (Frame-wise) & 1.3B & 24.9 & 5 & 84.26 & 85.25 & 80.30 \\
    Self-Forcing++~\cite{cui2025selfforcing++} & 1.3B & 24.9 & 5 & 83.11 & 83.79 & 80.37 \\
    \midrule
    Ours (15-3-3) - Best Performance   &  1.3B &  25.8 & 5 & \textbf{85.07} & \textbf{86.33} & 80.02 \\
    Ours - (7-7-7) - Fastest & 1.3B & 29.4 & 5 & 84.63 & 85.89 & 79.59 \\
    Ours (3-3-3-3-3-3-3) &  1.3B & 24.9 & 5 & 84.03 & 84.91 & 79.95  \\
    \midrule
    Ours (12-6-3) - Best Performance &  1.3B & 45.6 & 3 & 84.29 & 85.33 & 80.12 \\
    Ours - (7-7-7) - Fastest & 1.3B & 48.9 &  3 & 83.92 & 84.96 & 79.78 \\
    Ours (3-3-3-3-3-3-3) &  1.3B & 41.5 & 3 & 83.91 & 84.85 & 80.16  \\
    \bottomrule
    \end{tabular}
    }
    \caption{Comparisons of performance for 5s videos. *: We sample videos from the official checkpoint and test its performance. Here, the NFE of the causal distillation method contains $N$ steps for denoising and 1 step for caching.}
    \label{tbl:5svideo}
\end{table}

\section{Experiments}

\begin{figure*}[t]
    \centering
    \includegraphics[width=\linewidth]{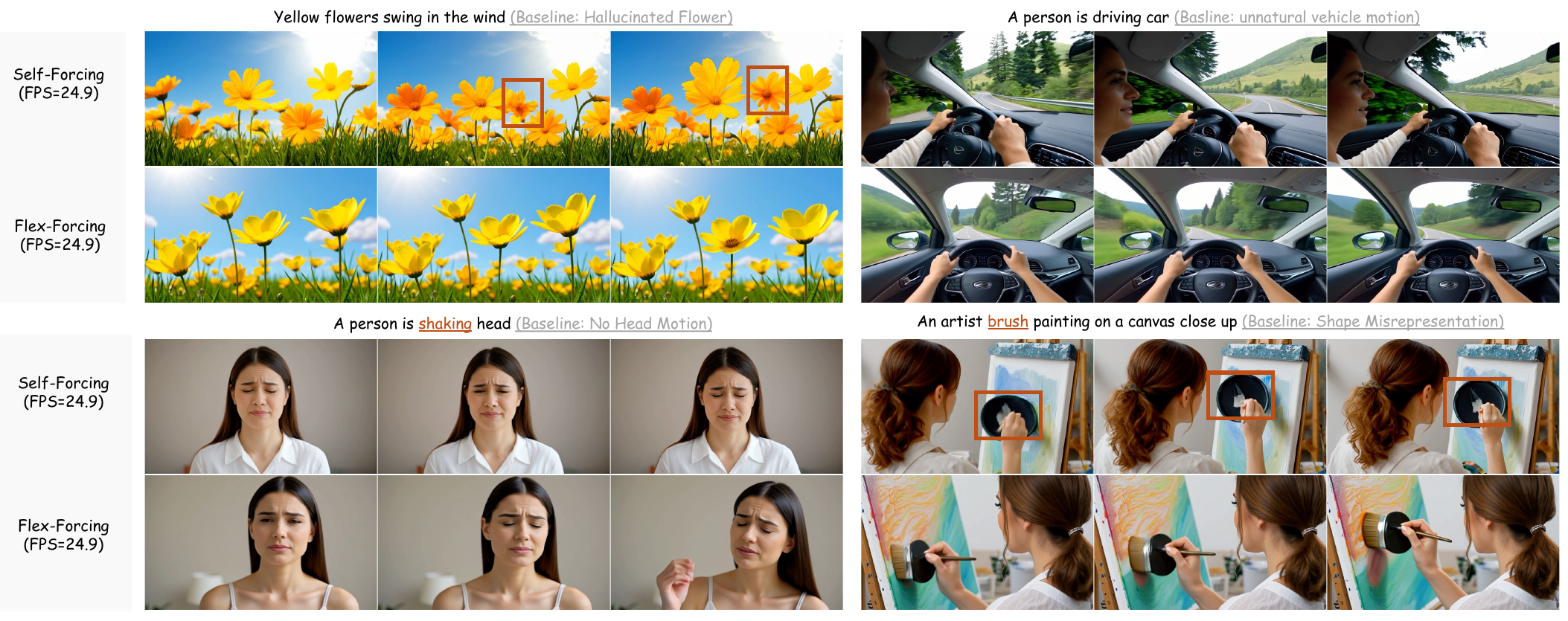}
    \caption{Visual comparison between Self-Forcing and Flex-Forcing in generating 5s videos. We show three frames at 0 s, 2.5 s, and 5 s. Flex-Forcing produces more coherent long-range motion and better alignment with the given instruction. }
    \label{fig:main_case_study}
    \vspace{-4mm}
\end{figure*}

\paragraph{Benchmark and baselines}

We build our baseline on Wan2.1-T2V-1.3B \cite{wan2025wan}, which generates 5-second videos at 832$\times$432 resolution. We evaluate inference efficiency and generation quality. Efficiency is measured in FPS for 81-frame generation on GB200. 
Quality is assessed using VBench \cite{huang2024vbench} for 5-second videos and VBench-Long \cite{huang2025vbench++} for 30-second videos. For fair comparison, we use the same prompts as Self-Forcing and fix random seeds across all experiments. We generate 5 samples per prompt for 5-second videos and one sample per prompt for 30-second videos.

We consider two groups of baselines. For the 5-second setting, we compare against Self-Forcing \cite{huang2025selfforcing} as well as recently proposed improvements built upon it \cite{cui2025selfforcing++}. Moreover, since our model also supports bidirectional diffusion inference, we additionally compare with few-step distillation methods for video diffusion models. For long-video evaluation, we compare our method with approaches that are train-short-evaluate-long \cite{huang2025selfforcing,yesiltepe2025infinity}.

\begin{table}[t]
    \centering
    \resizebox{\linewidth}{!}{%
    \begin{tabular}{c|c|ccc}
    \toprule
    Model & Steps &  \multicolumn{3}{c}{Evaluation Scores} \\
    & & Total & Quality & Semantic \\
    \midrule
    Wan2.1  & 50$\times$2& 84.26 & 85.30 & 80.09 \\ 
    \midrule
    DOLLAR \cite{ding2025dollar} & 4 & 82.57 & 83.83 & 77.51 \\ 
    T2V-Turbo-v2 \cite{li2024t2v} & 4 & 82.34 & 83.93 & 75.97 \\
    rCM \cite{zheng2025large}& 4 & 84.43 & 85.38 & 80.63 \\
    DMD-v \cite{nie2026transition}& 4 & 84.60 & 86.03 & 79.87 \\
    Flex-Forcing & 4 & 85.13 & 86.50 & 79.65 \\
    \midrule
    TMD \cite{nie2026transition} & 2.33 & 84.68 & 85.71 & 80.55 \\
    \midrule
    APT \cite{lin2025apt} & 2 & 81.85 & 84.39 & 71.70 \\
    rCM \cite{zheng2025large} & 2 & 84.09 & 84.90 & 80.86 \\
    DMD-v \cite{nie2026transition} & 2 & 84.39 & 85.65 & 79.32 \\
    Flex-Forcing & 2 & 84.20 & 85.27 & 79.92 \\
    \bottomrule
    \end{tabular}
    }
    \caption{Comparisons with few-step distilled diffusion models.}
    \label{tbl:5sdiffusion}
    \vspace{-5mm}
\end{table}

\begin{table*}[t]
    \centering
    \resizebox{\linewidth}{!}{%
    \begin{tabular}{c|cccccccc|cc}
    \toprule
    & Subject C. & Human Action & Aesthetic Quality & Temporal Style & Overall C. & Background C. & Appearance Style & Scene & \cellcolor{gray!20}{Quality Score} & \cellcolor{gray!20}{Semantic Score}\\
    \midrule
    LongLive$^*$ & 98.36 & 97.15 & 63.77 & 24.15 & 26.49 & 96.25 & 20.52 & 58.62 & 83.21 & 81.17\\
    \midrule
    Self-Forcing & 97.76 & 96.86 & 61.72 & 23.56 & 25.68 & 95.00 & 21.15 & 51.05 & 82.67 & 76.77 \\
    
    Infinity-RoPE & 97.68 & 96.99 & 62.33 & 23.74 & 26.26 & 96.00 & 20.91 & 53.47 & 83.77 & 79.11  \\
    Ours     & 98.00 \relativetext{ \shadowtext{(0.40$\uparrow$)}} & 97.07 \relativetext{ \shadowtext{(0.08$\uparrow$)}} & 63.31 \relativetext{ \shadowtext{(0.98$\uparrow$)}} & 24.26 \relativetext{ \shadowtext{(0.48$\uparrow$)}} & 25.92 \relativetext{ \shadowdowntext{(0.34$\downarrow$)}} & 95.50 \relativetext{ \shadowdowntext{(0.50$\downarrow$)}} & 20.32 \relativetext{ \shadowdowntext{(0.59$\downarrow$)}} & 52.20 \relativetext{ \shadowdowntext{(1.27$\downarrow$)}} & 85.30 & 78.86 \\
    \midrule
    & Imaging Quality & Color & Object Class & Flickering & Motion Smoothness & Dynamic Degree & Multiple Objects & Spatial & \cellcolor{gray!20}{Total Score} & \cellcolor{gray!20}{FPS} \\
    \midrule
    LongLive$^*$ & 68.35 & 90.07 & 94.08 & 99.36 & 98.81 & 38.89 & 85.86  & 82.06 & 82.80 & 25.07 \\
    \midrule
    Self-Forcing & 67.70 & 71.55 & 92.83 & 99.10 & 98.67 & 41.93 & 83.65 & 76.55 & 81.49 & 19.10 \\
    Infinity-RoPE & 69.30 & 82.52 & 91.90 & 99.28 & 98.72 & 50.26 & 83.55 & 82.48 & 82.84 & 19.10 \\
    Ours & 66.24 \relativetext{ \shadowdowntext{(3.06$\downarrow$)}} & 84.92 \relativetext{ \shadowtext{(2.40$\uparrow$)}} & 93.62 \relativetext{ \shadowtext{(1.72$\uparrow$)}} & 99.70 \relativetext{ \shadowtext{(0.42$\uparrow$)}} & 98.70 \relativetext{ \shadowdowntext{(0.02$\downarrow$)}} & 71.27 \relativetext{ \shadowtext{(21.01$\uparrow$)}} & 85.15 \relativetext{ \shadowtext{(1.60$\uparrow$)}} & 78.11 \relativetext{ \shadowdowntext{(4.37$\downarrow$)}} & 84.01 & 24.96 \\
    
    \bottomrule
    \end{tabular}
  }
    \caption{VBench-Long comparison on 30-second videos. $^*$Except for LongLive, all other methods are trained only on short videos.}
    \vspace{-6mm}
    \label{tab:30s_video}
\end{table*}

\paragraph{Implementations}
We follow Self-Forcing to use Wan2.1-T2V-1.3B as our base model and Wan2.1-T2V-14B as the teacher model to train Flex-Forcing. Training uses extended prompts from VidProM \cite{wang2024vidprom} for 600 iterations with a batch size of 64. We randomly pick the chunk size in training from 2 to 10. For the K-projection, we use a learning rate of 2e-6. For the 2-step model, we set the denoising steps set to $[1000, 500]$. The sink size for the long video is set to 3 latent frames, and the total attention window is 21 latent frames. 

\paragraph{Performance on 5s videos}
Results on 5-second videos are summarized in Tables~\ref{tbl:5svideo} and \ref{tbl:5sdiffusion}. For Flex-Forcing, we report three representative configurations: the best-performing setting, the speed-optimized setting, and the configuration matched to the baseline. Flex-Forcing consistently outperforms the baselines in both inference efficiency and video quality. We further compare Flex-Forcing with few-step distilled diffusion methods. Even in this regime, Flex-Forcing achieves comparable or superior video quality, indicating that the proposed framework preserves strong performance under fully bidirectional operation.
Qualitative case studies are presented in Figure~\ref{fig:main_case_study}. The primary improvements arise in temporal dynamics: long-horizon planning enables  coherent long-range motion, mitigates error accumulation, and yields smoother and more temporally consistent generation.

\paragraph{Performance of hybrid chunking over timesteps}

\begin{figure}[t]
    \centering
    \includegraphics[width=\linewidth]{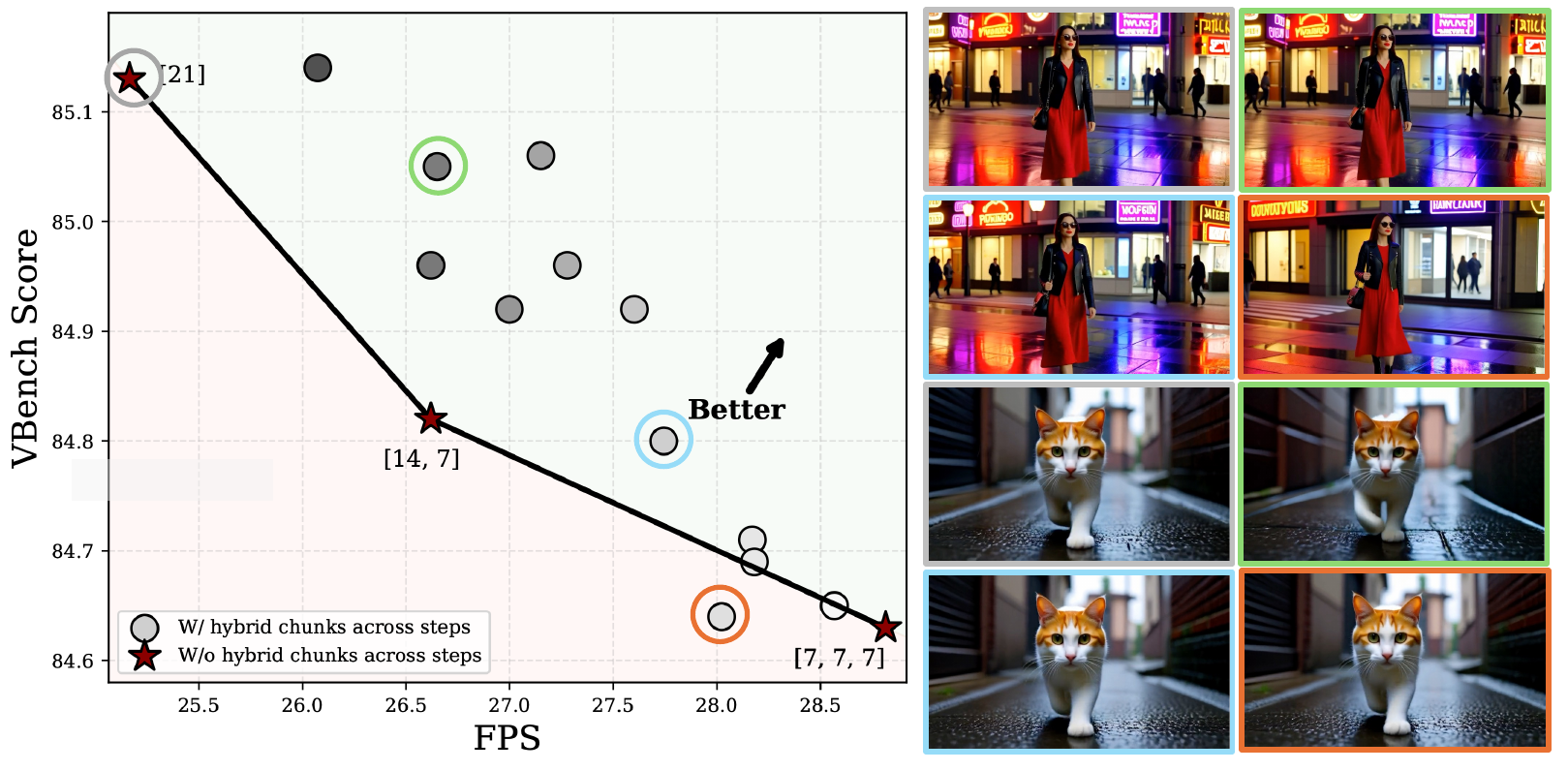}
    \caption{Results for hybrid chunking over denoising time. The points with lighter color represents those with those have more fine-grained splits.}
    \label{fig:hybrid}
    \vspace{-5mm}
\end{figure}

\begin{figure}[t]
    \centering
    \includegraphics[width=\linewidth]{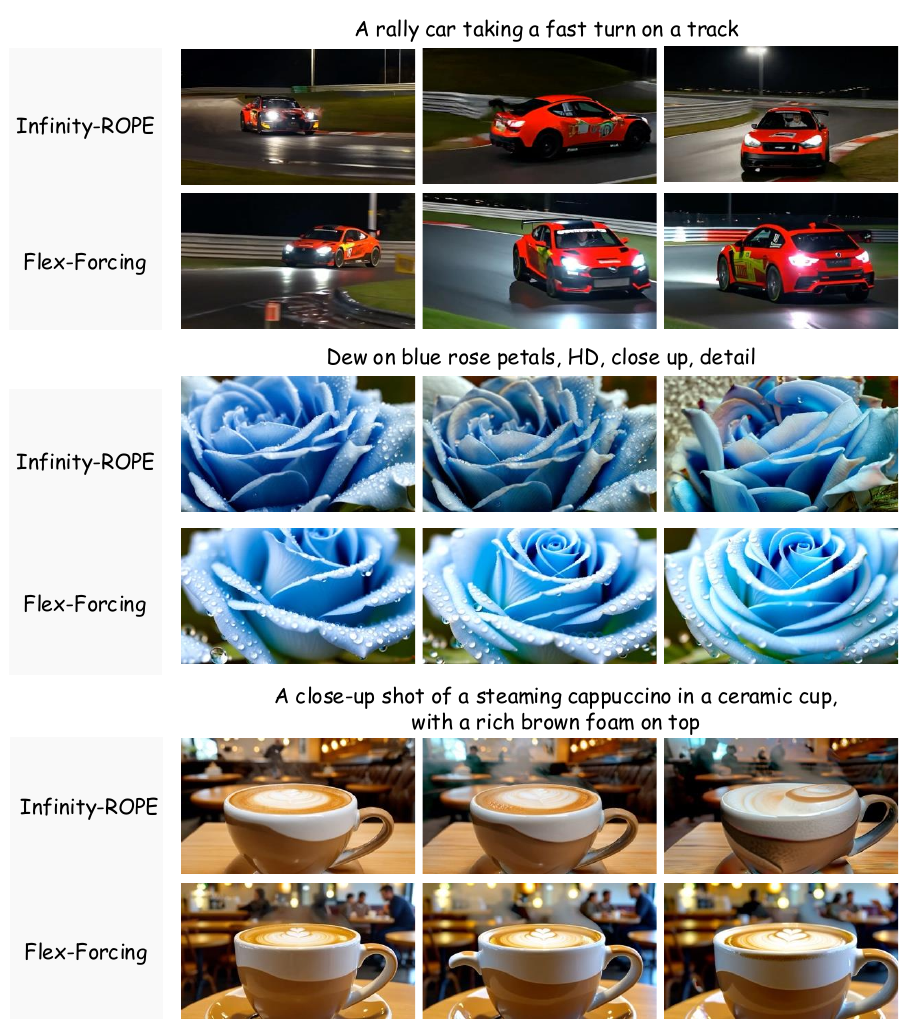}
    \caption{Comparisons between Flex-Forcing and Infinity-RoPE. Our method produces long video with improved semantic consistency, better aesthetic quality and mothion smoothness.}
    \label{fig:plan_w_diffusion}
\end{figure}

\begin{figure}[t]
    \centering
    \begin{subfigure}{0.49\linewidth}
        \centering
        \includegraphics[width=\linewidth]{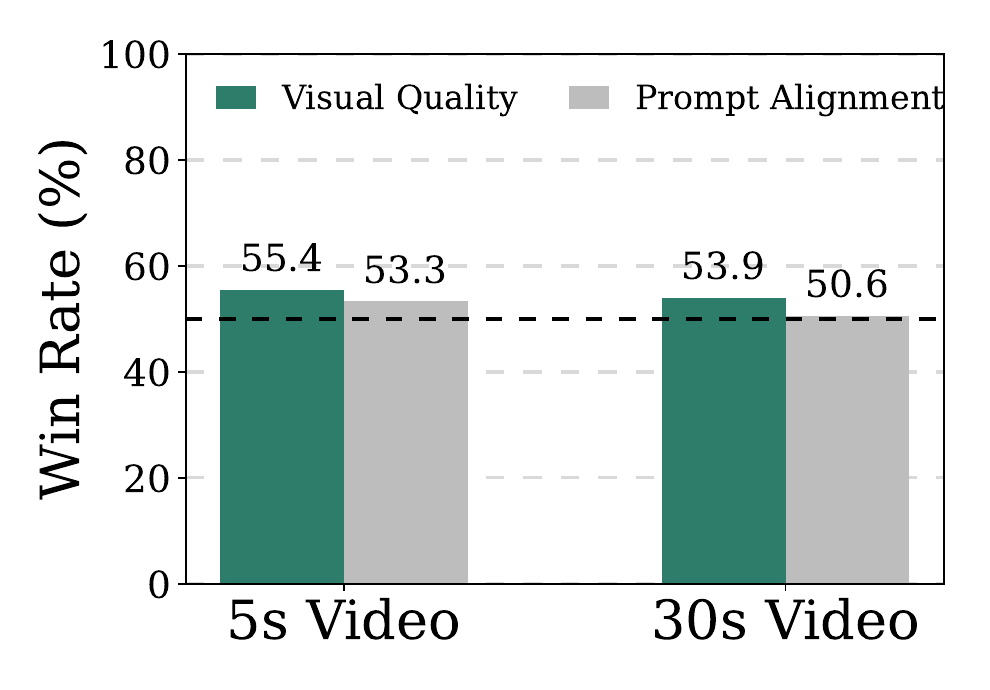}
        \label{fig:sub1}
    \end{subfigure}
    \hfill
    \begin{subfigure}{0.49\linewidth}
        \centering
        \includegraphics[width=\linewidth]{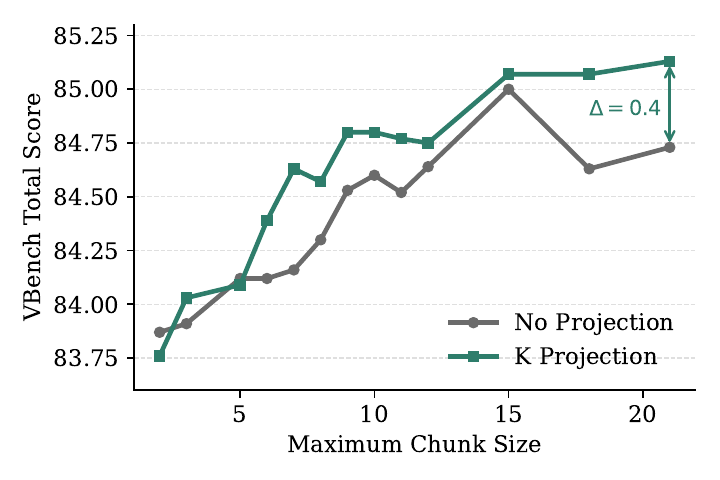}
        \label{fig:sub2}
    \end{subfigure}
    \vspace{-3mm}
    \caption{(Left) User study: We compare Flex-Forcing with Self-Forcing under 5s and 30s settings. (Right) Ablation of K-projection and its impact on different chunk configurations.}
    \label{fig:user_study}
\end{figure}

We further analyze the effect of hybrid chunking on denoising timesteps through an ablation study. Specifically, we consider three chunking configurations, 
$[21]$, $[14, 7]$ and $[7, 7, 7]$, where each configuration specifies the chunk sizes assigned to the denoising steps. We evaluate all settings under identical conditions, and the quantitative results are summarized in Figure~\ref{fig:hybrid}. Overall, hybrid chunking consistently achieves a more favorable trade-off between efficiency and quality compared to applying chunking only along temporal frame, effectively yielding a better Pareto frontier. We conduct qualitative analysis and observe that even the weakest-performing configuration, where the first denoising step uses a chunk size of 21 and the remaining stages use a chunk size of 7, still significantly outperforms Self-Forcing (84.3), while producing video content and fine-grained details similar to that of fully bidirectional model.

\paragraph{Performance on 30s videos}

Table \ref{tab:30s_video} shows the evaluation results of our method under the 30s setting. We build the inference of our method upon Infinity-RoPE ~\cite{yesiltepe2025infinity}. We observe that our method outperforms Infinity-RoPE across the majority of evaluation metrics, while also achieving faster inference speed. Compared to Infinity RoPE, our improvements primarily stem from enhanced video quality rather than semantic alignment, as our method does not explicitly optimize for alignment objectives. 
Notably, the most significant gains are observed in the dynamic degree, where our approach demonstrates a clear advantage. We further provide qualitative comparisons in the appendix to illustrate these improvements, showing substantially richer and more expressive motion dynamics, where the baseline tends to generate repetitive motions, resulting in reduced dynamic diversity.

\paragraph{User preference study} As shown in Figure \ref{fig:user_study}, our method are preferred than self-forcing both on the 5s and 30s setting. The advantage is more significant for the dimension of visual quality.

\paragraph{Ablation study: impact of K-projection}
We evaluate the impact of K-projection by comparing models trained with and without it under various inference configurations (Figure~\ref{fig:user_study}). Incorporating 
K-projection consistently improves performance across all settings, with the largest gains observed near the fully bidirectional regime. Without K-projection, performance degrades as chunk size increases; with K-projection, performance remains stable and improves smoothly. This demonstrates that K-projection stabilizes training when approaching bidirectional supervision and mitigates performance drops at large chunk sizes.

\section{Conclusion}
In summary, Flex-Forcing reframes bidirectional and autoregressive video generation not as competing paradigms, but a controllable inference dimension that can be smoothly adjusted according to requirements. By introducing the flexible chunking along the temporal and denoising axes, and bridging the different level of noise in causal and non-causal contexts, the model can adapt its causal structure at test time, enabling better globally coherent generation, efficient and better long-video synthesis and better trade-offs between quality and speed. 

\section*{Limitations}
Our method still has several limitations. Although it relaxes the strict left-to-right training constraint, the training–inference mismatch is not fully resolved and can still accumulate errors in long video generation. In addition, the effectiveness of the method appears to depend heavily on capabilities inherited from pre-training, especially bidirectional encoding priors, which may limit its transferability to models where such priors are weak or absent. More broadly, as with other generative video models, the proposed method may be misused in downstream applications if deployed without appropriate safeguards.

\section*{Impact Statement}
Flex-Forcing unifies bidirectional and autoregressive paradigms into a single model, reducing the computational cost and environmental footprint of maintaining separate specialized architectures. By allowing dynamic trade-offs between speed and quality at runtime, it democratizes high-fidelity, long-form video generation for resource-constrained environments. Additionally, the order-agnostic capabilities provide precise, localized editing tools for creative and industrial sectors, promoting more adaptive, resource-aware generative systems in media, simulation, and embodied AI.

\section*{Acknowledgement}
This research is in part supported by 
the Ministry of Education, Singapore, under the Academic Research Fund Tier~1 (FY2026, WBS: A-8004345-00-00).


\bibliography{example_paper}

@article{wan2025wan,
  title={Wan: Open and advanced large-scale video generative models},
  author={Wan, Team and Wang, Ang and Ai, Baole and Wen, Bin and Mao, Chaojie and Xie, Chen-Wei and Chen, Di and Yu, Feiwu and Zhao, Haiming and Yang, Jianxiao and others},
  journal={arXiv preprint arXiv:2503.20314},
  year={2025}
}

@article{huang2025selfforcing,
  title={Self Forcing: Bridging the Train-Test Gap in Autoregressive Video Diffusion},
  author={Huang, Xun and Li, Zhengqi and He, Guande and Zhou, Mingyuan and Shechtman, Eli},
  journal={arXiv preprint arXiv:2506.08009},
  year={2025}
}

@article{cui2025selfforcing++,
  title={Self-Forcing++: Towards Minute-Scale High-Quality Video Generation},
  author={Cui, Justin and Wu, Jie and Li, Ming and Yang, Tao and Li, Xiaojie and Wang, Rui and Bai, Andrew and Ban, Yuanhao and Hsieh, Cho-Jui},
  journal={arXiv preprint arXiv:2510.02283},
  year={2025}
}

@inproceedings{yin2025causvid,
  title={From slow bidirectional to fast autoregressive video diffusion models},
  author={Yin, Tianwei and Zhang, Qiang and Zhang, Richard and Freeman, William T and Durand, Fredo and Shechtman, Eli and Huang, Xun},
  booktitle={Proceedings of the Computer Vision and Pattern Recognition Conference},
  pages={22963--22974},
  year={2025}
}

@inproceedings{yin2024dmd,
  title={One-step diffusion with distribution matching distillation},
  author={Yin, Tianwei and Gharbi, Micha{\"e}l and Zhang, Richard and Shechtman, Eli and Durand, Fredo and Freeman, William T and Park, Taesung},
  booktitle={Proceedings of the IEEE/CVF conference on computer vision and pattern recognition},
  pages={6613--6623},
  year={2024}
}

@article{yin2024dmd2,
  title={Improved distribution matching distillation for fast image synthesis},
  author={Yin, Tianwei and Gharbi, Micha{\"e}l and Park, Taesung and Zhang, Richard and Shechtman, Eli and Durand, Fredo and Freeman, Bill},
  journal={Advances in neural information processing systems},
  volume={37},
  pages={47455--47487},
  year={2024}
}

@article{nie2026transition,
  title={Transition Matching Distillation for Fast Video Generation},
  author={Nie, Weili and Berner, Julius and Ma, Nanye and Liu, Chao and Xie, Saining and Vahdat, Arash},
  journal={arXiv preprint arXiv:2601.09881},
  year={2026}
}

@article{yesiltepe2025infinity,
  title={Infinity-RoPE: Action-Controllable Infinite Video Generation Emerges From Autoregressive Self-Rollout},
  author={Yesiltepe, Hidir and Meral, Tuna Han Salih and Akan, Adil Kaan and Oktay, Kaan and Yanardag, Pinar},
  journal={arXiv preprint arXiv:2511.20649},
  year={2025}
}

@inproceedings{huang2024vbench,
  title={Vbench: Comprehensive benchmark suite for video generative models},
  author={Huang, Ziqi and He, Yinan and Yu, Jiashuo and Zhang, Fan and Si, Chenyang and Jiang, Yuming and Zhang, Yuanhan and Wu, Tianxing and Jin, Qingyang and Chanpaisit, Nattapol and others},
  booktitle={Proceedings of the IEEE/CVF Conference on Computer Vision and Pattern Recognition},
  pages={21807--21818},
  year={2024}
}

@article{huang2025vbench++,
  title={Vbench++: Comprehensive and versatile benchmark suite for video generative models},
  author={Huang, Ziqi and Zhang, Fan and Xu, Xiaojie and He, Yinan and Yu, Jiashuo and Dong, Ziyue and Ma, Qianli and Chanpaisit, Nattapol and Si, Chenyang and Jiang, Yuming and others},
  journal={IEEE Transactions on Pattern Analysis and Machine Intelligence},
  year={2025},
  publisher={IEEE}
}

@inproceedings{ding2025dollar,
  title={Dollar: Few-step video generation via distillation and latent reward optimization},
  author={Ding, Zihan and Jin, Chi and Liu, Difan and Zheng, Haitian and Singh, Krishna Kumar and Zhang, Qiang and Kang, Yan and Lin, Zhe and Liu, Yuchen},
  booktitle={Proceedings of the IEEE/CVF International Conference on Computer Vision},
  pages={17961--17971},
  year={2025}
}

@article{li2024t2v,
  title={T2v-turbo-v2: Enhancing video generation model post-training through data, reward, and conditional guidance design},
  author={Li, Jiachen and Long, Qian and Zheng, Jian and Gao, Xiaofeng and Piramuthu, Robinson and Chen, Wenhu and Wang, William Yang},
  journal={arXiv preprint arXiv:2410.05677},
  year={2024}
}

@article{lin2025apt,
  title={Diffusion adversarial post-training for one-step video generation},
  author={Lin, Shanchuan and Xia, Xin and Ren, Yuxi and Yang, Ceyuan and Xiao, Xuefeng and Jiang, Lu},
  journal={arXiv preprint arXiv:2501.08316},
  year={2025}
}

@article{zheng2025large,
  title={Large scale diffusion distillation via score-regularized continuous-time consistency},
  author={Zheng, Kaiwen and Wang, Yuji and Ma, Qianli and Chen, Huayu and Zhang, Jintao and Balaji, Yogesh and Chen, Jianfei and Liu, Ming-Yu and Zhu, Jun and Zhang, Qinsheng},
  journal={arXiv preprint arXiv:2510.08431},
  year={2025}
}

@article{yang2025longlive,
  title={Longlive: Real-time interactive long video generation},
  author={Yang, Shuai and Huang, Wei and Chu, Ruihang and Xiao, Yicheng and Zhao, Yuyang and Wang, Xianbang and Li, Muyang and Xie, Enze and Chen, Yingcong and Lu, Yao and others},
  journal={arXiv preprint arXiv:2509.22622},
  year={2025}
}

@article{wang2024vidprom,
  title={Vidprom: A million-scale real prompt-gallery dataset for text-to-video diffusion models},
  author={Wang, Wenhao and Yang, Yi},
  journal={Advances in Neural Information Processing Systems},
  volume={37},
  pages={65618--65642},
  year={2024}
}

@article{chen2025seedance,
  title={Seedance 1.5 pro: A Native Audio-Visual Joint Generation Foundation Model},
  author={Chen, Siyan and Chen, Yanfei and Chen, Ying and Chen, Zhuo and Cheng, Feng and Chi, Xuyan and Cong, Jian and Cui, Qinpeng and Dong, Qide and Fan, Junliang and others},
  journal={arXiv preprint arXiv:2512.13507},
  year={2025}
}

@article{hacohen2026ltx,
  title={LTX-2: Efficient Joint Audio-Visual Foundation Model},
  author={HaCohen, Yoav and Brazowski, Benny and Chiprut, Nisan and Bitterman, Yaki and Kvochko, Andrew and Berkowitz, Avishai and Shalem, Daniel and Lifschitz, Daphna and Moshe, Dudu and Porat, Eitan and others},
  journal={arXiv preprint arXiv:2601.03233},
  year={2026}
}

@article{wu2025hunyuanvideo,
  title={Hunyuanvideo 1.5 technical report},
  author={Wu, Bing and Zou, Chang and Li, Changlin and Huang, Duojun and Yang, Fang and Tan, Hao and Peng, Jack and Wu, Jianbing and Xiong, Jiangfeng and Jiang, Jie and others},
  journal={arXiv preprint arXiv:2511.18870},
  year={2025}
}

@article{yang2024cogvideox,
  title={Cogvideox: Text-to-video diffusion models with an expert transformer},
  author={Yang, Zhuoyi and Teng, Jiayan and Zheng, Wendi and Ding, Ming and Huang, Shiyu and Xu, Jiazheng and Yang, Yuanming and Hong, Wenyi and Zhang, Xiaohan and Feng, Guanyu and others},
  journal={arXiv preprint arXiv:2408.06072},
  year={2024}
}

@article{chen2025skyreels,
  title={Skyreels-v2: Infinite-length film generative model},
  author={Chen, Guibin and Lin, Dixuan and Yang, Jiangping and Lin, Chunze and Zhu, Junchen and Fan, Mingyuan and Zhang, Hao and Chen, Sheng and Chen, Zheng and Ma, Chengcheng and others},
  journal={arXiv preprint arXiv:2504.13074},
  year={2025}
}

@article{he2025matrix,
  title={Matrix-game 2.0: An open-source real-time and streaming interactive world model},
  author={He, Xianglong and Peng, Chunli and Liu, Zexiang and Wang, Boyang and Zhang, Yifan and Cui, Qi and Kang, Fei and Jiang, Biao and An, Mengyin and Ren, Yangyang and others},
  journal={arXiv preprint arXiv:2508.13009},
  year={2025}
}

@article{hong2025relic,
  title={RELIC: Interactive Video World Model with Long-Horizon Memory},
  author={Hong, Yicong and Mei, Yiqun and Ge, Chongjian and Xu, Yiran and Zhou, Yang and Bi, Sai and Hold-Geoffroy, Yannick and Roberts, Mike and Fisher, Matthew and Shechtman, Eli and others},
  journal={arXiv preprint arXiv:2512.04040},
  year={2025}
}

@article{jiang2025vace,
  title={Vace: All-in-one video creation and editing},
  author={Jiang, Zeyinzi and Han, Zhen and Mao, Chaojie and Zhang, Jingfeng and Pan, Yulin and Liu, Yu},
  journal={arXiv preprint arXiv:2503.07598},
  year={2025}
}

@inproceedings{hu2024animate,
  title={Animate anyone: Consistent and controllable image-to-video synthesis for character animation},
  author={Hu, Li},
  booktitle={Proceedings of the IEEE/CVF Conference on Computer Vision and Pattern Recognition},
  pages={8153--8163},
  year={2024}
}

@article{deng2024nove,
  title={Autoregressive video generation without vector quantization},
  author={Deng, Haoge and Pan, Ting and Diao, Haiwen and Luo, Zhengxiong and Cui, Yufeng and Lu, Huchuan and Shan, Shiguang and Qi, Yonggang and Wang, Xinlong},
  journal={arXiv preprint arXiv:2412.14169},
  year={2024}
}

@article{teng2025magi,
  title={MAGI-1: Autoregressive Video Generation at Scale},
  author={Teng, Hansi and Jia, Hongyu and Sun, Lei and Li, Lingzhi and Li, Maolin and Tang, Mingqiu and Han, Shuai and Zhang, Tianning and Zhang, WQ and Luo, Weifeng and others},
  journal={arXiv preprint arXiv:2505.13211},
  year={2025}
}

@article{ho2022video,
  title={Video diffusion models},
  author={Ho, Jonathan and Salimans, Tim and Gritsenko, Alexey and Chan, William and Norouzi, Mohammad and Fleet, David J},
  journal={Advances in neural information processing systems},
  volume={35},
  pages={8633--8646},
  year={2022}
}

@inproceedings{yin2023nuwa,
  title={Nuwa-xl: Diffusion over diffusion for extremely long video generation},
  author={Yin, Shengming and Wu, Chenfei and Yang, Huan and Wang, Jianfeng and Wang, Xiaodong and Ni, Minheng and Yang, Zhengyuan and Li, Linjie and Liu, Shuguang and Yang, Fan and others},
  booktitle={Proceedings of the 61st Annual Meeting of the Association for Computational Linguistics (Volume 1: Long Papers)},
  pages={1309--1320},
  year={2023}
}

@article{singer2022make,
  title={Make-a-video: Text-to-video generation without text-video data},
  author={Singer, Uriel and Polyak, Adam and Hayes, Thomas and Yin, Xi and An, Jie and Zhang, Songyang and Hu, Qiyuan and Yang, Harry and Ashual, Oron and Gafni, Oran and others},
  journal={arXiv preprint arXiv:2209.14792},
  year={2022}
}

@inproceedings{yu2023magvit,
  title={Magvit: Masked generative video transformer},
  author={Yu, Lijun and Cheng, Yong and Sohn, Kihyuk and Lezama, Jos{\'e} and Zhang, Han and Chang, Huiwen and Hauptmann, Alexander G and Yang, Ming-Hsuan and Hao, Yuan and Essa, Irfan and others},
  booktitle={Proceedings of the IEEE/CVF Conference on Computer Vision and Pattern Recognition},
  pages={10459--10469},
  year={2023}
}

@article{hong2022cogvideo,
  title={Cogvideo: Large-scale pretraining for text-to-video generation via transformers},
  author={Hong, Wenyi and Ding, Ming and Zheng, Wendi and Liu, Xinghan and Tang, Jie},
  journal={arXiv preprint arXiv:2205.15868},
  year={2022}
}

@article{villegas2022phenaki,
  title={Phenaki: Variable length video generation from open domain textual description},
  author={Villegas, Ruben and Babaeizadeh, Mohammad and Kindermans, Pieter-Jan and Moraldo, Hernan and Zhang, Han and Saffar, Mohammad Taghi and Castro, Santiago and Kunze, Julius and Erhan, Dumitru},
  journal={arXiv preprint arXiv:2210.02399},
  year={2022}
}

@article{wang2024loong,
  title={Loong: Generating minute-level long videos with autoregressive language models},
  author={Wang, Yuqing and Xiong, Tianwei and Zhou, Daquan and Lin, Zhijie and Zhao, Yang and Kang, Bingyi and Feng, Jiashi and Liu, Xihui},
  journal={arXiv preprint arXiv:2410.02757},
  year={2024}
}

@inproceedings{chen2024videocrafter2,
  title={Videocrafter2: Overcoming data limitations for high-quality video diffusion models},
  author={Chen, Haoxin and Zhang, Yong and Cun, Xiaodong and Xia, Menghan and Wang, Xintao and Weng, Chao and Shan, Ying},
  booktitle={Proceedings of the IEEE/CVF Conference on Computer Vision and Pattern Recognition},
  pages={7310--7320},
  year={2024}
}

@article{liu2025rolling,
  title={Rolling forcing: Autoregressive long video diffusion in real time},
  author={Liu, Kunhao and Hu, Wenbo and Xu, Jiale and Shan, Ying and Lu, Shijian},
  journal={arXiv preprint arXiv:2509.25161},
  year={2025}
}

@article{yan2021videogpt,
  title={Videogpt: Video generation using vq-vae and transformers},
  author={Yan, Wilson and Zhang, Yunzhi and Abbeel, Pieter and Srinivas, Aravind},
  journal={arXiv preprint arXiv:2104.10157},
  year={2021}
}

@article{jin2024pyramidal,
  title={Pyramidal flow matching for efficient video generative modeling},
  author={Jin, Yang and Sun, Zhicheng and Li, Ningyuan and Xu, Kun and Jiang, Hao and Zhuang, Nan and Huang, Quzhe and Song, Yang and Mu, Yadong and Lin, Zhouchen},
  journal={arXiv preprint arXiv:2410.05954},
  year={2024}
}

@misc{openai2024sora,
  title={Sora},
  author={OpenAI},
  howpublished={\url{https://openai.com/index/sora}},
  year={2024}
}

@misc{kuaishou2024kling,
  title={Kling},
  author={Kuaishou},
  howpublished={\url{https://kling.kuaishou.com}},
  year={2024}
}

@misc{runway2024gen,
  title={Gen-3 Alpha},
  author={Runway},
  howpublished={\url{https://runwayml.com/research/introducing-gen-3-alpha}},
  year={2024}
}

@misc{google2024veo,
  title={Veo},
  author={Google},
  howpublished={\url{https://deepmind.google/technologies/veo/}},
  year={2024}
}

@article{yuan2025lumos,
  title={Lumos-1: On autoregressive video generation from a unified model perspective},
  author={Yuan, Hangjie and Chen, Weihua and Cen, Jun and Yu, Hu and Liang, Jingyun and Chang, Shuning and Lin, Zhihui and Feng, Tao and Liu, Pengwei and Xing, Jiazheng and others},
  journal={arXiv preprint arXiv:2507.08801},
  year={2025}
}

@article{chen2024diffusion,
  title={Diffusion forcing: Next-token prediction meets full-sequence diffusion},
  author={Chen, Boyuan and Mart{\'\i} Mons{\'o}, Diego and Du, Yilun and Simchowitz, Max and Tedrake, Russ and Sitzmann, Vincent},
  journal={Advances in Neural Information Processing Systems},
  volume={37},
  pages={24081--24125},
  year={2024}
}

@article{blattmann2023stable,
  title={Stable video diffusion: Scaling latent video diffusion models to large datasets},
  author={Blattmann, Andreas and Dockhorn, Tim and Kulal, Sumith and Mendelevitch, Daniel and Kilian, Maciej and Lorenz, Dominik and Levi, Yam and English, Zion and Voleti, Vikram and Letts, Adam and others},
  journal={arXiv preprint arXiv:2311.15127},
  year={2023}
}

@inproceedings{blattmann2023align,
  title={Align your latents: High-resolution video synthesis with latent diffusion models},
  author={Blattmann, Andreas and Rombach, Robin and Ling, Huan and Dockhorn, Tim and Kim, Seung Wook and Fidler, Sanja and Kreis, Karsten},
  booktitle={Proceedings of the IEEE/CVF conference on computer vision and pattern recognition},
  pages={22563--22575},
  year={2023}
}

@inproceedings{gupta2024photorealistic,
  title={Photorealistic video generation with diffusion models},
  author={Gupta, Agrim and Yu, Lijun and Sohn, Kihyuk and Gu, Xiuye and Hahn, Meera and Li, Fei-Fei and Essa, Irfan and Jiang, Lu and Lezama, Jos{\'e}},
  booktitle={European Conference on Computer Vision},
  pages={393--411},
  year={2024},
  organization={Springer}
}

@article{ho2022imagen,
  title={Imagen video: High definition video generation with diffusion models},
  author={Ho, Jonathan and Chan, William and Saharia, Chitwan and Whang, Jay and Gao, Ruiqi and Gritsenko, Alexey and Kingma, Diederik P and Poole, Ben and Norouzi, Mohammad and Fleet, David J and others},
  journal={arXiv preprint arXiv:2210.02303},
  year={2022}
}

@inproceedings{bar2024lumiere,
  title={Lumiere: A space-time diffusion model for video generation},
  author={Bar-Tal, Omer and Chefer, Hila and Tov, Omer and Herrmann, Charles and Paiss, Roni and Zada, Shiran and Ephrat, Ariel and Hur, Junhwa and Liu, Guanghui and Raj, Amit and others},
  booktitle={SIGGRAPH Asia 2024 Conference Papers},
  pages={1--11},
  year={2024}
}

@article{luo2023videofusion,
  title={Videofusion: Decomposed diffusion models for high-quality video generation},
  author={Luo, Zhengxiong and Chen, Dayou and Zhang, Yingya and Huang, Yan and Wang, Liang and Shen, Yujun and Zhao, Deli and Zhou, Jingren and Tan, Tieniu},
  journal={arXiv preprint arXiv:2303.08320},
  year={2023}
}

@inproceedings{dhariwal2021diffusion,
  title={Diffusion Models Beat {GAN}s on Image Synthesis},
  author={Dhariwal, Prafulla and Nichol, Alexander},
  booktitle=NIPS,
  pages={8780--8794},
  year={2021}
}

@inproceedings{rombach2022high,
  title={High-Resolution Image Synthesis with Latent Diffusion Models},
  author={Rombach, Robin and Blattmann, Andreas and Lorenz, Dominik and Esser, Patrick and Ommer, Bj{\"o}rn},
  booktitle=CVPR,
  pages={10684--10695},
  year={2022}
}

@inproceedings{karras2022elucidating,
  title={Elucidating the Design Space of Diffusion-Based Generative Models},
  author={Karras, Tero and Aittala, Miika and Aila, Timo and Laine, Samuli},
  booktitle=NIPS,
  pages={26565--26577},
  year={2022}
}

@misc{black2024flux,
  author={{Black Forest Labs}},
  title={{FLUX}},
  howpublished={\url{https://github.com/black-forest-labs/flux}},
  year=2024
}

@inproceedings{esser2024scaling,
  title={Scaling rectified flow transformers for high-resolution image synthesis},
  author={Esser, Patrick and Kulal, Sumith and Blattmann, Andreas and Entezari, Rahim and M{\"u}ller, Jonas and Saini, Harry and Levi, Yam and Lorenz, Dominik and Sauer, Axel and Boesel, Frederic and others},
  booktitle={Forty-first international conference on machine learning},
  year={2024}
}

@inproceedings{peebles2023scalable,
  title={Scalable diffusion models with transformers},
  author={Peebles, William and Xie, Saining},
  booktitle={Proceedings of the IEEE/CVF international conference on computer vision},
  pages={4195--4205},
  year={2023}
}
\bibliographystyle{icml2026}

\newpage
\appendix
\onecolumn

\section{Qualitative results on flexible chunks on video frames.}

\begin{figure}[htbp!]
    \centering
    \includegraphics[width=\linewidth]{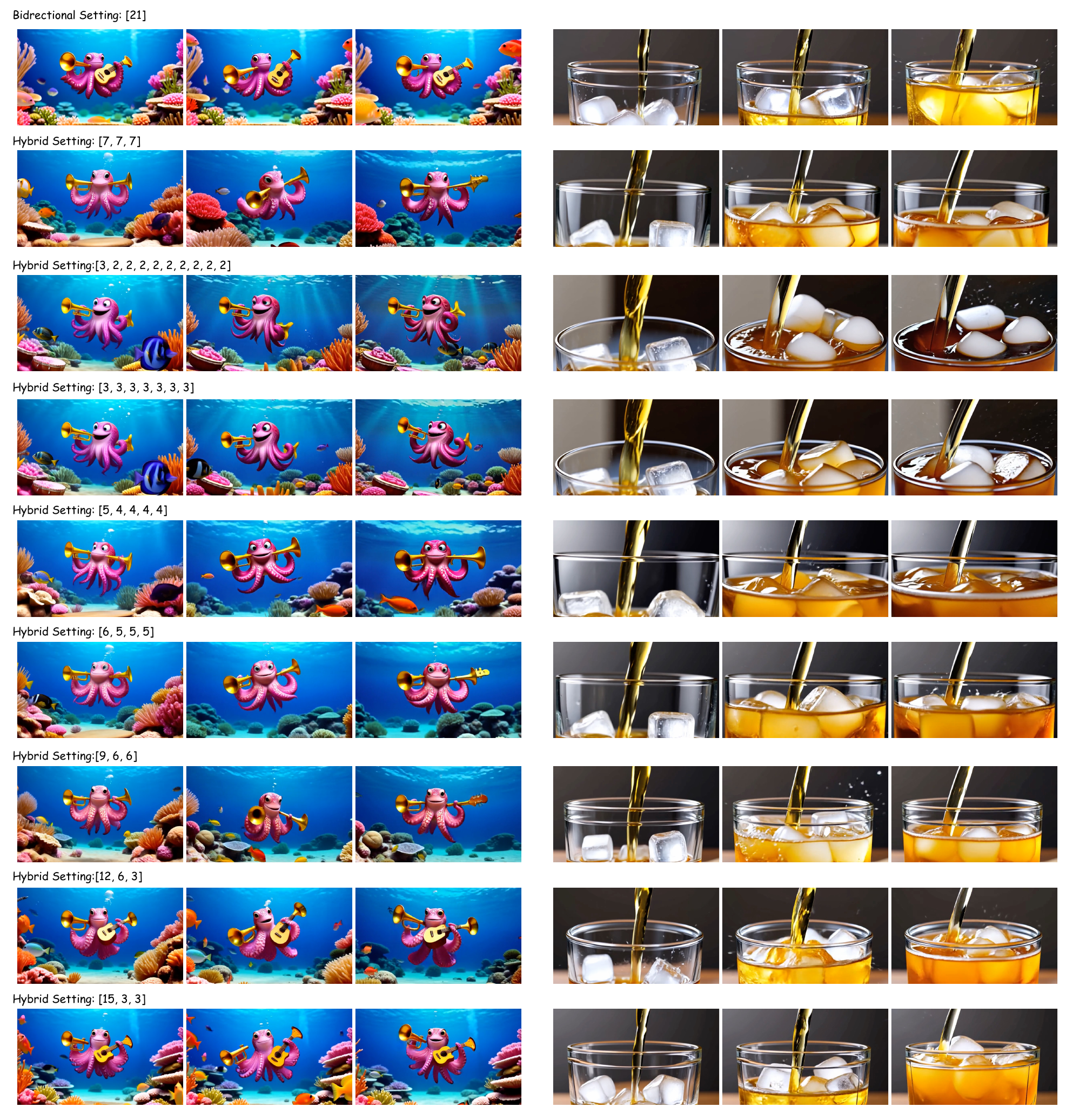}
    \caption{\textbf{Prompt (left):} A vibrant and lively underwater scene featuring an octopus playing multiple musical instruments simultaneously in a colorful band. The octopus has a playful and joyful expression, its tentacles deftly manipulating a trumpet, a drum, and a guitar. Its body is adorned with iridescent patterns, and it appears to be having fun. The background showcases a diverse array of marine life, including colorful fish and coral reefs, with a gentle underwater current flowing. The scene is captured in a dynamic angle, emphasizing the octopus's movements and the instruments it plays. The water has a soft, shimmering quality, enhancing the underwater atmosphere. A mid-shot with a dynamic camera angle. \textbf{Prompt (right)}: A close-up shot of a drink being poured over ice, showcasing the detailed flow of liquid interacting with the ice cubes. The drink cascades down, creating ripples and splashes on the surface of the ice, which glistens under the soft lighting. The glass holds a clear, amber-colored liquid, and the ice cubes sparkle with tiny droplets of condensation. The background is blurred, highlighting the dynamic interaction between the drink and the ice. The photo has a crisp, natural lighting style, emphasizing the fluid motion and the sparkling ice. A close-up from a slightly downward angle.}
    \label{fig:apx_frame}
\end{figure}

In the experiment that shown in Figure 1, we use the following 14 configurations: [21], [18,3], [15,3,3], [12,6,3], [11,10], [10,11], [9,9,3], [8,13], [8,8,5], [7,7,7], [9,6,6], [6,5,5,5], [5,4,4,4], [3,3,3,3,3,3,3], [3,2,2,2,2,2,2,2,2,2]. 
We here show some results when we set the chunking configurations to those difference choices in Figure.\ref{fig:apx_frame}. In these experiments, we use the same model with an identical random seed and vary only the chunking configuration. The results reveal a clear and consistent trend: as the average chunk size increases, the visual quality of the generated videos improves steadily and progressively approaches that of full bidirectional attention. In contrast, when the chunk size is small, we observe pronounced exposure bias, indicating that exposure bias still remains a significant issue under fine-grained chunking.

\begin{figure}[t]
    \centering
    \begin{subfigure}{0.49\linewidth}
        \centering
        \includegraphics[width=\linewidth]{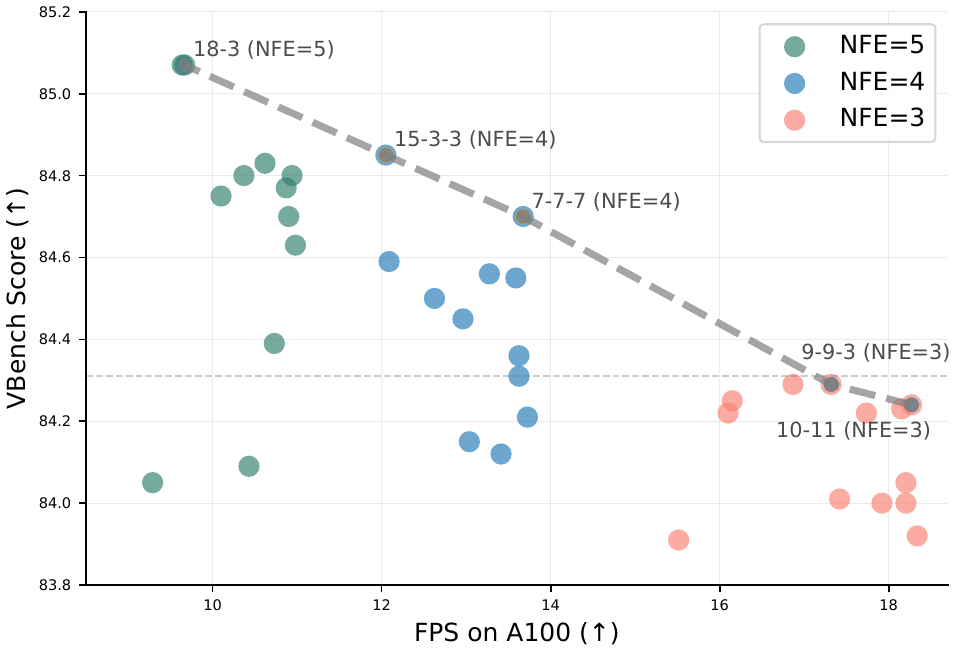}
        \label{fig:appendix_speed_a100}
    \end{subfigure}
    \hfill
    \begin{subfigure}{0.49\linewidth}
        \centering
        \includegraphics[width=\linewidth]{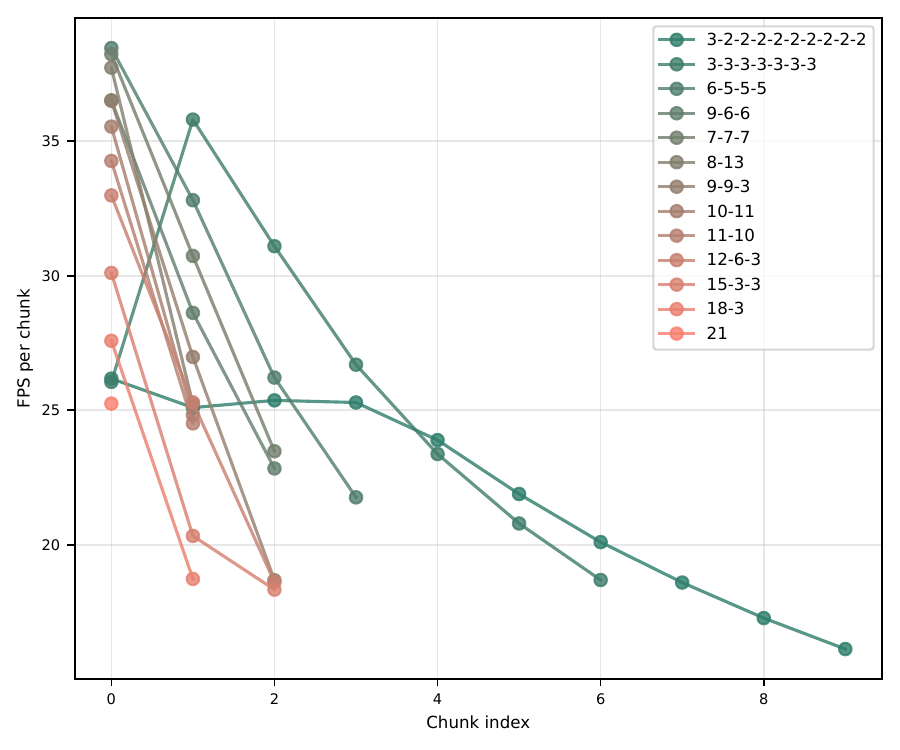}
        \label{fig:appendix_speed_per_chunj}
    \end{subfigure}
    \caption{(Left) The VBench score and FPS on a A100 GPU. We adopt the same configurations as on GB200. (Right) We plot the FPS in each chunk if we have different chunk configurations.}
    \label{fig:appendix_speed}
\end{figure}

\paragraph{Speed Analysis}
We further provide a detailed speed analysis. All experiments are conducted on a single A100 GPU, and the results are reported in Figure~\ref{fig:appendix_speed}. 
We then show the results if we calculate the FPS for each block. While smaller chunk sizes achieve higher per-chunk FPS, they require a larger number of rollout rounds, resulting in a lower overall throughput. In contrast, although each chunk operates at a relatively lower FPS, the total number of rollout rounds is substantially reduced compared to Self-Forcing, leading to higher end-to-end generation speed.

\section{Qualitative results on flexible chunks on denoising timesteps.}

\begin{figure}[h!]
    \centering
    \begin{minipage}[t]{0.48\textwidth}
        \vspace{0pt}
        
        Figure~\ref{fig:chunking} illustrates the processing order when applying flexible chunking over denoising timesteps. The 0-th block is bidirectional and thus does not depend on any prior KV cache, allowing immediate execution. In contrast, although Block 4 has access to intermediate representations from Block 1, it cannot be computed until Block 3 finishes, since the required KV caches from earlier frames are unavailable.
        
        We further study the impact of flexible chunking across denoising steps under multiple configurations. As shown in Figure~\ref{fig:appendix_timestep}, we evaluate three chunking schemes—[21], [14, 7], and [7, 7, 7]. All configurations yield results highly similar to the unchunked bidirectional setting, with differences mainly in local visual details. As chunking becomes more fine-grained, deviations from the bidirectional baseline become more evident, especially in later frames, suggesting reduced temporal consistency toward the end of the video.
    \end{minipage}
    \hfill
    \begin{minipage}[t]{0.48\textwidth}
        \vspace{0pt}
        \centering
        \includegraphics[width=\linewidth]{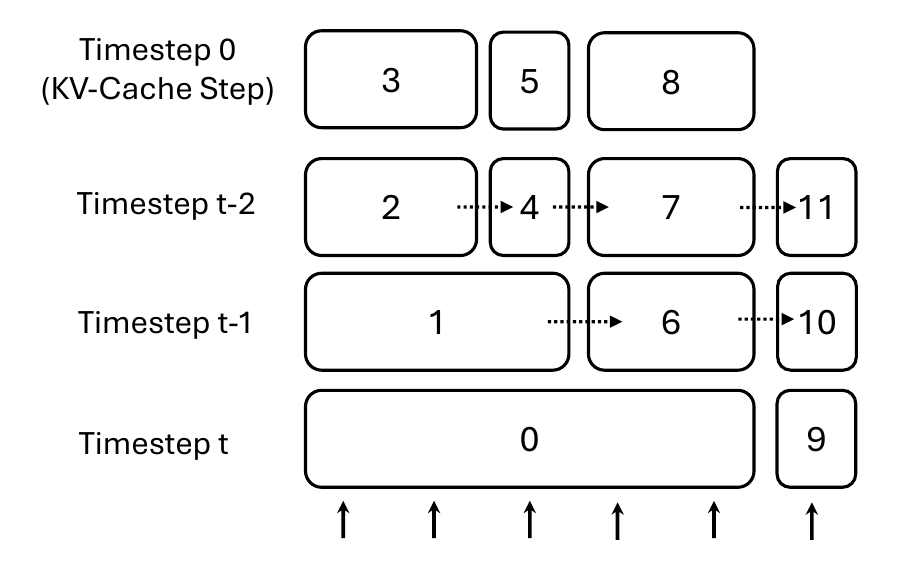}
        \caption{Illustration of flexible chunking across timesteps. The number on the block is the order they would be executed.}
        \label{fig:chunking}
    \end{minipage}
\end{figure}

\begin{figure}[tbp]
    \centering
    \includegraphics[width=\linewidth]{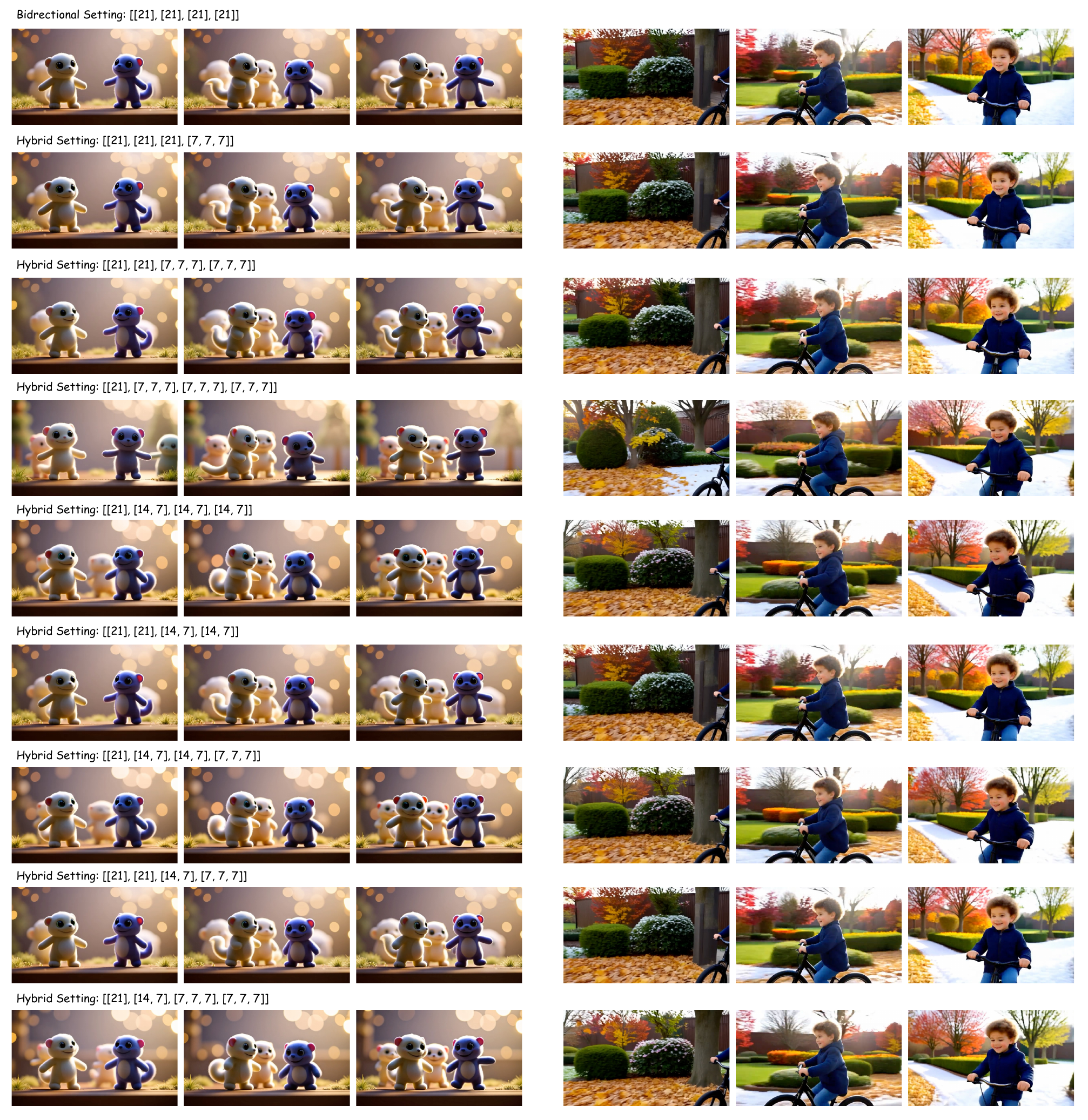}
    \caption{\textbf{Prompt (left): } A miniature 3D render in an octane engine style depicting adorable wool and felt monsters dancing together in a dreamy, bokeh-filled setting. These soft, cuddly creatures, with big expressive eyes and fluffy bodies, are illuminated by gentle, diffused lighting that casts a warm, ethereal glow. The background features a soft, hazy backdrop with a dreamy bokeh effect, adding a cinematic quality to the scene. The monsters are shown from various angles, capturing their playful movements and expressions, creating a charming and enchanting atmosphere. A medium shot with a dynamic camera angle, highlighting the natural and joyful dance of these woolen monsters. \textbf{Prompt (right):} A dynamic photograph capturing a little boy riding his bike through a garden that transitions through the changing seasons—fall leaves crunch underfoot, winter snow blankets the ground, spring flowers bloom, and summer sunshine sparkles through the foliage. The boy, with curly brown hair and a joyful smile, pedals energetically, his arms outstretched in excitement. The garden backdrop features trees with branches adorned in each season’s distinctive foliage. A series of shots taken from various angles, starting with a wide shot of the boy entering the garden in spring, transitioning to a mid-shot of him biking through the colorful autumn leaves, then a close-up of him riding through a snowy path, and finally a wide-angle view of him enjoying the warm summer sun. The photo has a natural, documentary style, emphasizing the boy’s natural movements and the vibrant colors of the changing seasons.}
    \label{fig:appendix_timestep}
\end{figure}

\section{Case Study}
\begin{figure}[tbp]
    \centering
    \includegraphics[width=\linewidth]{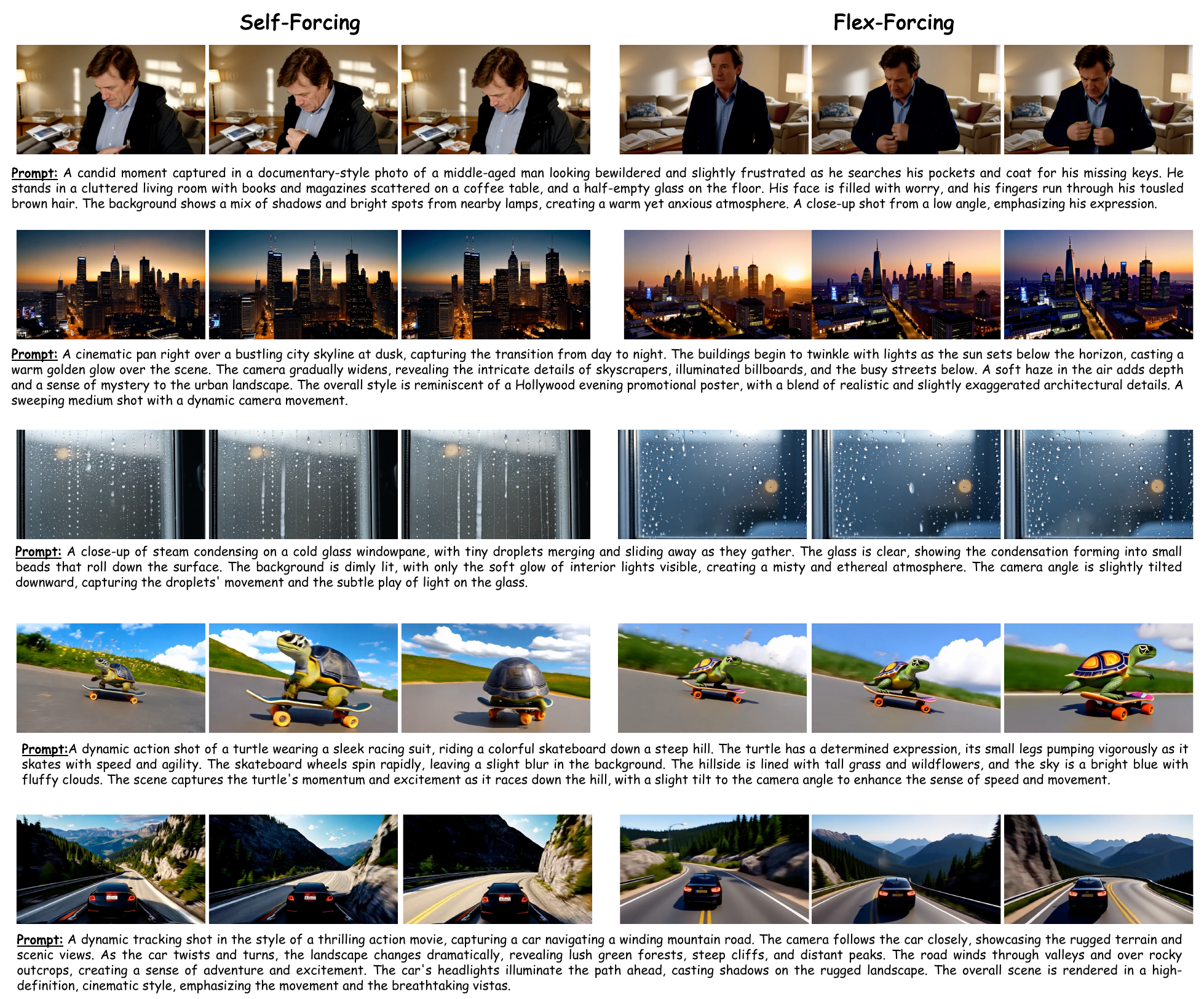}
    \caption{Visual Comparisons between Flex-Forcing and Self-Forcing.}
    \label{fig:appendix_case_study}
\end{figure}

\begin{figure}[t!]
    \centering
    \includegraphics[width=\linewidth]{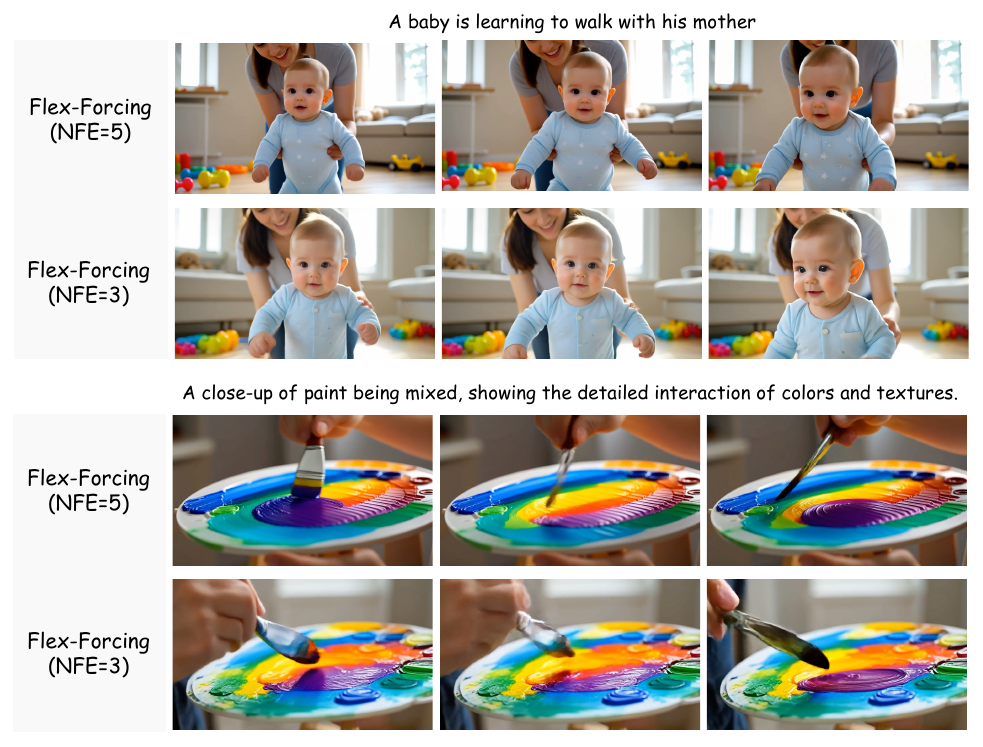}
    \caption{Visual Comparisons between Flex-Forcing with NFE=5 and NFE=3.  }
    \label{fig:appendix_2step}
\end{figure}

we provide further comparisons between Self-Forcing and Flex-Forcing to complement the main results (See Figure \ref{fig:appendix_case_study}). We also compare Flex-Forcing under different numbers of function evaluations with NFE=5 and NFE=3 (See Figure \ref{fig:appendix_2step}), to analyze the impact of denoising steps on generation quality and efficiency. 

\section{Impact Statement}
This work advances the foundations of video generation by demonstrating that bidirectional diffusion and autoregressive generation, traditionally treated as mutually exclusive paradigms, can be unified within a single, flexible model. By enabling dynamic trade-offs between generation quality, efficiency, and controllability at inference time, the proposed framework reduces the need to train and maintain multiple specialized models, thereby lowering computational cost and environmental footprint. Beyond improving scalability for long-form video generation, this flexibility unlocks new applications such as localized, order-agnostic video editing and efficient refinement under constrained budgets. We anticipate that these capabilities will benefit a wide range of downstream domains, including content creation, simulation, and embodied AI, while encouraging future research toward adaptive, resource-aware generative systems that better align model capabilities with real-world constraints.

\end{document}